\documentclass{article}
\usepackage[T1]{fontenc}
\usepackage[utf8]{inputenc}
\usepackage{bm}
\usepackage{amsmath}
\usepackage{graphicx}
\usepackage{microtype}
\usepackage{hyperref}
\hypersetup{unicode=true,
 breaklinks=false,pdfborder={0 0 1},colorlinks=false}

\makeatletter
\usepackage[numbers]{natbib}

\usepackage[final]{neurips_2022}

\usepackage{url}
\usepackage{amsfonts}
\usepackage{nicefrac}
\usepackage{xcolor}
\usepackage{sidecap}

\title{Globally Gated Deep Linear Networks}
  
  \author{%
    Qianyi Li$^{1}$ \quad Haim Sompolinsky$^{2,3}$\\
    $^1$Biophysics Graduate Program, Harvard University \quad \\
    $^2$Center for Brain Science, Harvard University \\
    \quad $^3$Edmond and Lily Safra Center for Brain Sciences, Hebrew University\\
    \texttt{qianyi\_li@g.harvard.edu},\quad
    \texttt{hsompolinsky@mcb.harvard.edu},\quad
    \texttt{haim@fiz.huji.ac.il}\\
}

\makeatother

\begin{document}

\maketitle

\begin{abstract}
Recently proposed Gated Linear Networks (GLNs) present a tractable
nonlinear network architecture, and exhibit interesting capabilities
such as learning with local error signals and reduced forgetting in
sequential learning. In this work, we introduce a novel gating architecture,
named Globally Gated Deep Linear Networks (GGDLNs) where gating units
are shared among all processing units in each layer, thereby decoupling
the architectures of the nonlinear but unlearned gating and the learned
linear processing motifs. We derive exact equations for the generalization
properties of Bayesian Learning in these networks in the finite-width
thermodynamic limit, defined by $N,P\rightarrow\infty$ while $P/N=O(1)$
where $N$ and $P$ are the hidden layers' width and size of training
data sets respectfully. We find that the statistics of the network
predictor can be expressed in terms of kernels that undergo shape
renormalization through a data-dependent order parameter matrix compared
to the infinite-width Gaussian Process (GP) kernels. Our theory accurately
captures the behavior of finite width GGDLNs trained with gradient
descent (GD) dynamics. We show that kernel shape renormalization gives
rise to rich generalization properties w.r.t. network width, depth
and $L_{2}$ regularization amplitude. Interestingly, networks with
a large number of gating units behave similarly to standard ReLU architectures.
Although gating units in the model do not participate in supervised
learning, we show the utility of unsupervised learning of the gating
parameters. Additionally, our theory allows the evaluation of the
network's ability for learning multiple tasks by incorporating task-relevant
information into the gating units. In summary, our work is the first
exact theoretical solution of learning in a family of nonlinear networks
with finite width. The rich and diverse behavior of the GGDLNs suggests
that they are helpful analytically tractable models of learning single
and multiple tasks, in finite-width nonlinear deep networks. 
\end{abstract}

\section{Introduction}

Despite the recent advances in machine learning, theoretical understanding
of how machine learning algorithms work is very limited. Many current
theoretical approaches study  infinitely
wide networks \cite{cho2009kernel,lee2017deep,matthews2018gaussian},
where the input-output relation is equivalent to a Gaussian Process
(GP) in function space with a covariance matrix defined by a GP kernel.
However, this GP limit holds when the network width approaches infinity
while the size of the training data remains finite, severely limiting
its applicability to realistic conditions. Another line of work focuses
on finite-width deep linear neural networks (DLNNs)\cite{li2021statistical,saxe2019mathematical,lampinen2018analytic},
while applicable in a wider regime, the generalization behavior of
linear networks are very limited, and the bias contribution always
remains constant with network parameters \cite{li2021statistical},
which fails to capture the behavior of generalization performance
in general nonlinear networks. Therefore, a tractable nonlinear network
architecture is in need for theoretically probing into the diverse
generalization behavior of general nonlinear networks.

Recently proposed Gated Linear Networks (GLNs) present a tractable
nonlinear network architecture \cite{budden2020gaussian,sezener2021rapid,veness2019gated},
with capabilities such as learning with local error signals and mitigating
catastrophic forgetting in sequential learning. Inspired by these
recent advances in GLNs, we propose Globally Gated Deep Linear Networks
(GGDLNs) as a simplified GLN structure that preserves the nonlinear
property of general GLNs, the decoupling of fixed nonlinear gating
from learned linear processing units, and the ability to separate the
processing of multiple tasks using the gating units. Our GGDLN structure
is different from previous GLNs in several ways. First, the gating
units are shared across hidden layer units and different layers while
in previous work each unit has its own set of gatings \cite{fiat2019decoupling,sezener2021rapid,veness2019gated,saxe2022neural}.
Second, we define global learning objective instead of local errors
\cite{sezener2021rapid,veness2019gated}. These simplifications allow
us to obtain direct analytical expressions of memory capacity and
exact generalization error of these networks for arbitrary training
and testing data, providing quantitative insight into the effect of
learning in nonlinear networks, as opposed to previous studies of generalization
bounds \cite{fiat2019decoupling}, expressivity estimates \cite{veness2019gated,veness2017online}, learning dynamics with very specific assumptions on data structure \cite{saxe2022neural}
and indirect quantities relevant to generalization such as the implicit
bias of the network \cite{lippl2022implicit}. Furthermore, the kernel
expression of the predictor statistics we propose in this work also
allow us to make qualitative explanations of the generalization and
how it's related to data structure and network representation for
single and multiple tasks.

First, we introduce the architecture of our GGDLNs and analyze its
memory capacity. We then derive our theory for generalization properties
of GGDLNs, and make qualitative connections between the generalization
behavior and the relation between the renormalization matrix and task
structure. Second, we apply our theory to GGDLNs performing multiple
tasks, focusing on two scenarios where tasks are either defined by different
input statistics or different output labels on the same inputs.
While the effect of kernel renormalization is different in the two
cases, we find that for fixed gating functions, de-correlation between tasks
always improves generalization.

\section{\label{sec:Globally-gated-deep}Globally gated deep linear networks}

\begin{figure}[h]
\centering

\includegraphics[width=.95\textwidth]{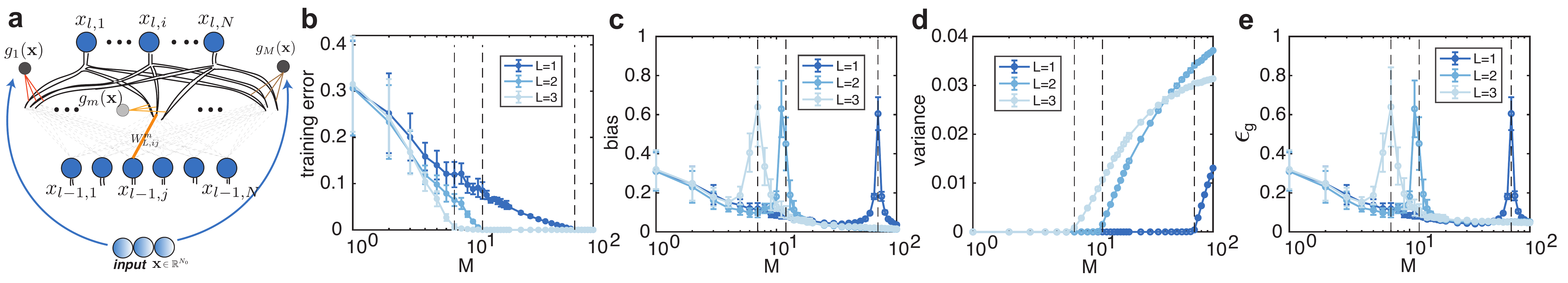}\caption{\label{fig:Globally-gated-deep}Globally gated deep linear networks.
(\textbf{a}) Structure of GGDLNs, each neuron in the hidden layer has
$M$ dendrites, each with a different input-dependent gating $g_{m}({\bf x})$
which is fixed during training, the $M$ gatings are shared across
neurons in the hidden layer. The $m$-th dendritic branch of the $i$-th
neuron in layer $l$ connects to neuron $j$ in the previous layer
with weight $W_{l,ij}^{m}$(shown in orange). (\textbf{b}) Training error of networks with 1-3 hidden layers in the GP limit as a function of $M$ evaluated on a noisy ReLU teacher task, training error goes to zero at network capacity (black dashed lines). (\textbf{c}-\textbf{e})
Bias, variance and generalization error of the same network and task as (b). Bias and generalization error diverges, variance generalization becomes nonzero
at network capacity (black dashed line). See Appendix $\text{\ref{SI:noisyrelu}}$
for detailed parameters.}
\end{figure}

In GGDLNs, the network input-output relation is defined as follows,
\begin{equation}
f({\bf x})=\frac{1}{\sqrt{NM}}\sum_{i=1}^{N}\sum_{m=1}^{M}a_{m,i}x_{L,i}g_{m}({\bf x}),\ x_{l,i}=\begin{cases}
\frac{1}{\sqrt{N_{0}M}}\sum_{j=1}^{N_{0}}\sum_{m=1}^{M}W_{l,ij}^{m}g_{m}({\bf x})x_{l-1,j} & l>1\\
\frac{1}{\sqrt{N_{0}}}\sum_{j=1}^{N_{0}}W_{l,ij}x_{l-1,j} & l=1
\end{cases}\label{eq:input-output}
\end{equation}

where ${\bf x}_{0}={\bf x}$ is the input, $N$ is the hidden layer
width, $M$ is the number of gating units in each layer, and $N_{0}$
is the input dimension. Each neuron in every layer has $M$ dendrites,
each with an input-dependent global gating $g_{m}({\bf x})$ shared
across all neurons. The $m$-th dendritic branch of neuron $i$ in
the $L$-th hidden layer connects to neurons in the previous layer
with a dendrite-specific weight vector ${\bf W}_{L,i}^{m}$ (or with
readout weight vector ${\bf a}_{m}$ for the output neuron), as shown
in Fig.$\text{\ref{fig:Globally-gated-deep} (a)}$. Note that although
the gatings are fixed during learning, changes in the weights affect
how these gatings act on the hidden layer activations, and it is
interesting to understand how the learned task interacts with these
gating operations. Since adding gatings at the input layer is equivalent
to expanding the input dimension and replacing $x_{j}$ by $x_{j}g_{m}({\bf x})$,
and learning does not affect how the gatings interact with the input,
\emph{we do not add gatings at the input layer for simplicity. }

\textbf{Memory Capacity: }Memory capacity refers to the maximum number
of random (or generic) input-output examples for which there exists
a set of parameters such that the network achieves zero training error
(here we consider the mean squared error, MSE). By definition, it
is irrespective of the learning algorithm. The capacity bounds of
deep nonlinear networks has been extensively studied in many recent
works \cite{fiat2019decoupling,vershynin2020memory,yamasaki1993lower}.
To calculate the capacity of GGDLNs, note that the input-output relation
given by Eq.$\text{\ref{eq:input-output} }$ can be alternatively
expressed as $f({\bf x})=\sum_{m_{1},\cdots,m_{L},j}W_{m_{1},\cdots,m_{L},j}^{{\rm eff}}x_{m_{1},\cdots,m_{L},j}^{{\rm eff}}$,
which is a linear combination of the effective input $x_{m_{1},\cdots,m_{L},j}^{{\rm eff}}=g_{m_{1}}({\bf x})g_{m_{2}}({\bf x})\cdots g_{m_{L}}({\bf x})x_{j}$
($m_{l}=1,\cdots,M$;$j=1,\cdots,N_{0}$), with some effective weights
${\bf W}^{{\rm eff}}$ which is a complicated function of ${\bf a}$
and ${\bf W}$. Here $m_{l}$ is the index of the gatings in the $l$-th
layer. As the gating units are shared across layers, the effective
input ${\bf x}^{{\rm eff}}$ has $N_{0}\binom{M+L-1}{L}$ independent
dimensions. This combinatorial term represents the number of possible combinations of L gatings selected from M total number of gatings. Assuming $N\gg M^{L}$ such that the effective weight $W_{m_{1},\cdots,m_{L},j}^{{\rm eff}}$
can take any desired value in the $N_{0}M^{L}$ dimensional whole space, the problem of finding ${\bf W}^{{\rm eff}}$ with zero
training error is equivalent to a linear regression problem with input
${\bf x}^{{\rm eff}}$ and the target outputs. Therefore, the capacity
is equivalent to the number of independent input dimensions, given
by $P\leq N_{0}\binom{M+L-1}{L}$.\\
The above capacity is verified by Fig.$\text{\ref{fig:Globally-gated-deep}}$(b), where the training error becomes nonzero above the memory capacity. The generalization behavior also changes drastically
at network capacity (Fig.$\text{\ref{fig:Globally-gated-deep}}$(c-e)), where generalization error and its bias contribution
diverge, and the variance contribution shrinks to 0 (see detailed
calculation in the next paragraph and Appendix $\text{\ref{subsec:Generalization}}$).
This double descent property of the generalization error is similar
to previously studied in linear and nonlinear networks. Furthermore,
although the output of the network is a linear function of the effective
input ${\bf x}^{{\rm eff}}$, due to the multiplicative nature of
the network weights and the gatings, learning in GGDLNs is highly
nonlinear and the space of solution for ${\bf W}$ and ${\bf a}$
is highly nontrivial, and the network exhibit properties unique to
nonlinear networks, as we will show in the following sections. 

\textbf{Posterior distribution of network weights:} We consider a Bayesian network setup, where
the network weights are random variables whose statistics are determined
by the training data and network parameters, instead of deterministic
variables. This probabilistic approach enables us to study the properties
of \textit{the entire solution space} instead of a single solution which may
be heavily initialization dependent. We consider the posterior distribution
of the network weights induced by learning with a Gaussian prior \cite{amit1987statistical,advani2013statistical,bahri2020statistical,engel2001statistical},
given by 
\begin{equation}
P({\bf \Theta})=Z^{-1}\exp(-\frac{1}{2T}\sum_{\mu=1}^{P}(f({\bf x}^{\mu},{\bf \Theta})-Y^{\mu})^{2}-\frac{1}{2\sigma^{2}}{\bf \Theta}^{\top}{\bf \Theta})\label{eq:posterior}
\end{equation}
where $Z$ is the partition function $Z=\int d{\bf \Theta}P({\bf \Theta})$.
The first term in the exponent is the MSE of the
network outputs on a set of $P$ training data points ${\bf x}^{\mu}$
from their target outputs $Y^{\mu}$, and the second term is a Gaussian
prior on the network parameters $\Theta=\{{\bf W},{\bf a}\}$ with
amplitude $\sigma^{-2}.$ In this work we focus on the $T\rightarrow0$
limit where the first term dominates. Below the network
capacity, the distribution of $\Theta$ concentrates onto the solution space that yields zero training error, the Gaussian prior then
biases the solution space towards weights with smaller $L_{2}$ norms.
The fundamental properties of the system can be derived from the partition
function. As the distribution is quadratic in the readout weights
$a_{m,i}$, it is straightforward to integrate them out, which yields
\begin{equation}
Z=\int d{\bf W}\exp[-\frac{1}{2\sigma^{2}}\mathrm{Tr}({\bf W}^{\top}{\bf W})+\frac{1}{2}{\bf Y}^{\top}{\bf K}_{L}({\bf W})^{-1}{\bf Y}+\frac{1}{2}\log\det({\bf K}_{L}({\bf W}))]\label{eq:W-dependent}
\end{equation}
where $\text{{\bf W}}$ denotes all the remaining weights in the network,
and ${\bf K}_{L}({\bf W})$ is the ${\bf W}$ dependent $P\times P$
kernel on the training data, defined as $K_{L}^{\mu\nu}({\bf W})=(\frac{\sigma^{2}}{M}{\bf g}({\bf x}^{\mu})^{\top}{\bf g}({\bf x}^{\nu}))(\frac{1}{N}{\bf x}_{L}^{\mu}({\bf W})^{\top}{\bf x}_{L}^{\nu}({\bf W}))$. 

\textbf{Generalization in infinitely wide GGDLNs: }It is well known
that in infinitely wide networks where $N\rightarrow\infty$ while
$P$ remains finite (also referred to as the GP limit), ${\bf K}_{L}({\bf W})$
is self-averaging and does not depend on the specific realization
of ${\bf W}$. It can therefore be replaced by the GP kernel defined
as $\langle{\bf K}_{L}({\bf W})\rangle_{{\bf W}}$ where ${\bf W}\sim\mathcal{N}(0,\sigma^{2})$
\cite{lee2017deep}. For GGDLNs, the GP kernel for a pair of arbitrary
data ${\bf x}$ and ${\bf y}$ is given by $K_{GP}({\bf x},{\bf y})=(\frac{\sigma^{2}}{M}{\bf g}({\bf x})^{\top}{\bf g}({\bf y}))^{L}K_{0}({\bf x},{\bf y})$,
where $K_{0}({\bf x},{\bf y})=\frac{\sigma^{2}}{N_{0}}{\bf x}^{\top}{\bf y}$.
We denote the $P\times P$ kernel data $matrix$ as ${\bf K}_{GP}$
where $K_{GP}^{\mu\nu}=K_{GP}({\bf x}^{\mu},{\bf x}^{\nu})$, and
the input kernel matrix on training data as ${\bf K}_{0}$ where $K_{0}^{\mu\nu}=K_{0}({\bf x}^{\mu},{\bf x}^{\nu}).$

Generalization error is measured by MSE including the bias and the variance contributions,
$\epsilon_{g}=\underbrace{(\langle f({\bf x})\rangle_{{\bf \Theta}}-y({\bf x}))^{2}}_{\text{bias}}+\underbrace{\langle\delta f({\bf x})^{2}\rangle_{\Theta}}_{\text{variance}}$,
which depends on the first and second order statistics of the predictor.
In the GP limit, we have 
\begin{equation}\label{eq:gp}
\langle f({\bf x})\rangle={\bf k}_{GP}({\bf x})^{\top}{\bf K}_{GP}^{-1}{\bf Y},\ \langle\delta f({\bf x})^{2}\rangle={\bf K}_{GP}({\bf x},{\bf x})-{\bf k}_{GP}({\bf x})^{\top}{\bf K}_{GP}^{-1}{\bf k}_{GP}({\bf x})
\end{equation}
where $k_{GP}^{\mu}({\bf x})=K_{GP}({\bf x},{\bf x}^{\mu})$. Note
that the rank of ${\bf K}_{GP}$ is the same as the capacity of the
network, and the kernel matrix becomes singular as $P$ approaches
its capacity (the interpolation threshold), which results in nonzero training error, diverging
bias and vanishing variance contribution to the generalization error (Fig.$\text{\ref{fig:Globally-gated-deep} }$(b-e)).
The singularity of the kernel at the interpolation threshold holds
also for finite width networks, and similar diverging bias and vanishing
variance are seen in our finite width theory below (Section $\text{\ref{sec:Kernel-shape-renormalization} }$)
and are confirmed by simulation of networks trained with GD (see Appendix
$\text{\ref{subsec:Double-descent-with}}$,\cite{lecun-mnisthandwrittendigit-2010}). 

\section{\label{sec:Kernel-shape-renormalization}Kernel shape renormalization
theory in finite-width GLNs}

We now address the finite width thermodynamic limit, where $P,N\rightarrow\infty$
but $P/N\sim\mathcal{O}(1)$, $M,L\sim\mathcal{O}(1)$. In this limit,
calculating the statistics of the network predictor requires integration
over ${\bf W}$ in Eq.$\text{\ref{eq:W-dependent} }$. To do so, we
apply the previous method of Back-propagating Kernel Renormalization
\cite{li2021statistical} (see Appendix $\text{\ref{sec:Derivation}}$)
to GGDLNs. The partition function for a single hidden layer network
is given by $Z=\exp(-H_{1})$, where the Hamiltonian $H_{1}$ is given
by 
\begin{align}\label{eq:hamiltonian}
H_{1} & =\frac{1}{2}{\bf Y}^{\top}\tilde{{\bf K}}_{1}^{-1}{\bf Y}+\frac{1}{2}\log\det(\tilde{{\bf K}}_{1})-\frac{N}{2}\log\det{\bf U}_{1}+\frac{1}{2\sigma^{2}}N\mathrm{Tr}({\bf U}_{1})\nonumber \\
\tilde{K}_{1}^{\mu\nu} & =(\frac{1}{M}{\bf g}({\bf x}^{\mu})^{\top}{\bf U}_{1}{\bf g}({\bf x}^{\nu}))K_{0}^{\mu\nu}
\end{align}
Comparing the matrix $\tilde{{\bf K}}_{1}$ to ${\bf K}_{GP}$, we
note that the GP kernel is renormalized by an an $M\times M$ matrix
order parameter ${\bf U}_{1}$. This order paramter satisfies the
self-consistent equation 
\begin{equation}
{\bf U}_{1}=I-\frac{1}{NM}{\bf U}_{1}^{1/2}{\bf g}^{\top}[\tilde{{\bf K}}_{1}^{-1}\circ{\bf K}_{0}]{\bf g}{\bf U}_{1}^{1/2}+\frac{1}{NM}{\bf U}_{1}^{1/2}{\bf g}^{\top}[\tilde{{\bf K}}_{1}^{-1}{\bf Y}{\bf Y}^{\top}\tilde{{\bf K}}_{1}^{-1}\circ{\bf K}_{0}]{\bf g}{\bf U}_{1}^{1/2}\label{eq:spequation}
\end{equation}
where $\circ$ denotes element-wise multiplication. In the linear
case (which corresponds to $M=1)$, the GP kernel is renormalized
by a scalar factor. In the $M>1$ case, the effect of renormalization
is more drastic as it changes that not only the amplitude but also
the shape of the kernel. The renormalization matrix has an interesting
physical interpretation that relates it to the readout weights ${\bf a}$
of GGDLNs, 
\begin{equation}
U_{1}^{mn}=\langle\frac{1}{N}\sum_{i=1}^{N}a_{m,i}a_{n,i}\rangle\label{eq:interpretation}
\end{equation}
The calculation can be extended to multiple layers with a new order
parameter introduced for each layer (see Appendix $\text{\ref{sec:Derivation}}$). The predictor statistics for a input ${\bf x}$ can be expressed in
terms of the renormalized kernels, for a network with $L=1$
\begin{equation}
\langle f({\bf x})\rangle_{{\bf \Theta}}=\tilde{{\bf k}}_{1}({\bf x})^{\top}\tilde{{\bf K}}_{1}^{-1}{\bf Y},\langle\delta f({\bf x})^{2}\rangle_{{\bf \Theta}}=\tilde{K}_{1}({\bf x},{\bf x})-\tilde{{\bf k}}_{1}({\bf x})^{\top}\tilde{{\bf K}}_{1}^{-1}\tilde{{\bf k}}_{1}({\bf x})\label{eq:predictorstatistics}
\end{equation}
where $\tilde{K}_{1}({\bf x},{\bf y})=(\frac{1}{M}{\bf g}({\bf x})^{\top}{\bf U}_{1}{\bf g}({\bf y}))K_{0}({\bf x},{\bf y})$
denotes the renormalized kernel \textit{function} for two arbitrary inputs ${\bf x}$
and ${\bf y}$, $\tilde{{\bf K}}_{1}$ denotes the $P\times P$ renormalized
kernel \textit{matrix} on the training data, and $\tilde{{\bf k}}_{1}({\bf x})$ is
a $P$-dimensional vector, $\tilde{k}_{1}^\mu({\bf x})=K({\bf x},{\bf x}^{\mu})$.
The kernel renormalization in GGDLNs changes the shape of the kernel
through the data dependent ${\bf U}_{1}$, reflecting the nonlinear
property of the network, and resulting in more complex behavior of
predictor statistics relative to the linear networks, as shown in
Section $\text{\ref{sec:Generalization}}$. Our theory describes the
properties of the posterior distribution of the network weights induced
at equilibrium by Langevin dynamics with the MSE cost function and
the Gaussian prior \cite{li2021statistical,naveh2021predicting,zinn2021quantum,risken1989fpe}.
Simulating this dynamics agrees remarkably well with the simulation
(see Appendix $\text{\ref{subsec:Langevin-dynamics}}$). Although
our theoretical results do not directly describe the solutions obtained
by running gradient descent (GD) dynamics on the training
error, it is interesting to ask to what extent the predicted behaviors of
our theory are also exhibited by GD dynamics of the same network architectures, as GD-based learning is more widely used.
We will compare our theoretical results with numerics of GD dynamics throughout the paper. We consider the case where
the network is initialized with Gaussian i.i.d. weights with variance
$\sigma^{2}$, and the mean and variance of the predictors are evaluated
across multiple initializations (see Appendix $\text{\ref{subsec:Gradient-descent-numerics} }$for
details). As we will show, our theory makes accurate qualitative predictions
for GD dynamics in all examples in this paper, in the sense that while the exact values may not match, the general trend of how generalization or representation varies with different parameters in different regimes are very similar. 

\vspace{-0.2cm}
\section{Generalization \label{sec:Generalization}}

For linear networks the generalization error depends on $N,\sigma^{2}$
and $L$ through the variance only, while the mean predictor always
assumes the same value as in the GP limit \cite{li2021statistical}.
This is because the scalar kernel renormalization of $\tilde{{\bf k}}_{1}({\bf x})$
is cancelled out in the mean predictor by the renormalization of the
inverse kernel $\tilde{{\bf K}}_{1}^{-1}$. In contrast, for GGDLNs
the mean predictor and hence the error bias also change with these
network parameters due to the matrix nature of the kernel renormalization (Eq.$\text{\ref{eq:predictorstatistics}}$)
. Below we investigate in detail how matrix renormalization of the
kernel affects the generalization behavior (especially the bias term)
of the network.

\subsection{Networks with single hidden layer}

\begin{figure}[h]
\centering

\includegraphics[width=0.95\textwidth]{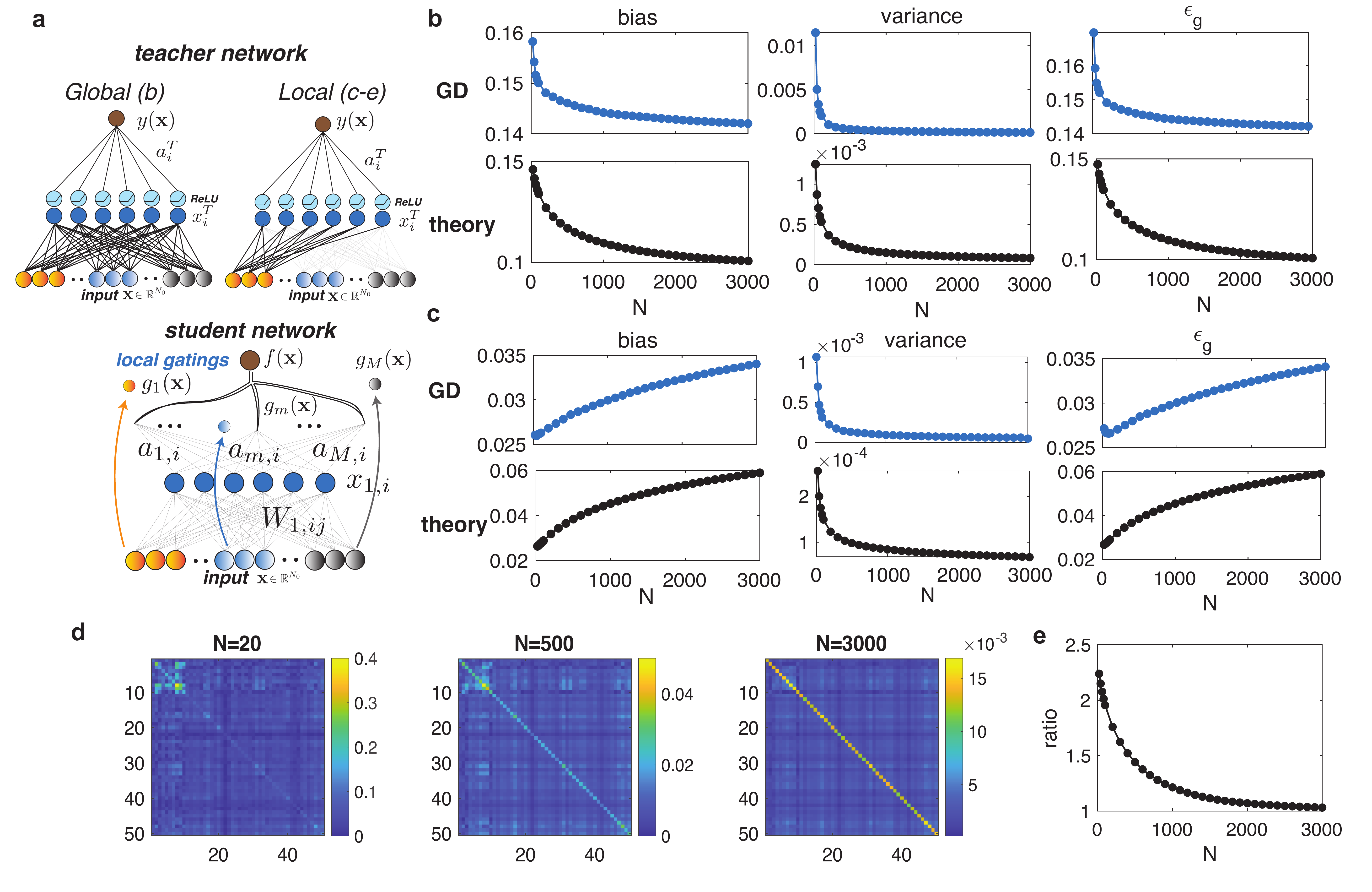}\caption{\label{fig:Dependence-of-generalization}Dependence of generalization
error on network width for a ReLU teacher task. (\textbf{a})\textit{Top:}
The ReLU teacher network, the input ${\bf x}$ is divided into 5 subsets
of input dimensions, the input layer weights either assume same order
of magnitude across different input dimensions (left, (b)), or assume
larger amplitudes for one subset of input dimensions-the preferred
inputs\textbf{ }(right, bold connections to a subset of input neurons,(c-e)).
\textit{Bottom:} The student network is a GGDLN with one hidden layer
and gatings with localized receptive fields: each gating is connected
to only a subset of input dimensions.\textbf{ }(\textbf{b}) Bias,
variance and generalization error decreases as a function of $N$
for a regular ReLU teacher, theory agrees qualitatively well with
GD dynamics. (\textbf{c}) Bias and generalization error increases
as a function of $N$ for ReLU teacher with preferred inputs. (\textbf{d})
Amplitude of the renormalization matrix ${\bf U}_{1}$ for different network widths
for the teacher with preferred inputs. The first $10\times10$ block
corresponds to the gatings with the same receptive field as the teacher's
preferred inputs, and is amplified for small $N$. (\textbf{e}) The
ratio of the average amplitude of the first $10\times10$ block relative
to the average amplitude of the other four $10\times10$ diagonal
blocks decreases as a function of $N$.}
\end{figure}

\textbf{Feature selection in finite-width networks:} Unlike in DLNs,
the bias term in GGDLNs depends on $N$, exhibiting different dependence
in different parameter regimes. This dependence also varies with the
choice of the gating functions. In Fig.$\text{\ref{fig:Dependence-of-generalization}}$ we
consider a student-teacher learning task, commonly used for evaluating
and understanding neural network performance\cite{watanabe2017student,wang2021knowledge,advani2020high,seung1992statistical}.
We present results of learning a ReLU teacher task in GGDLNs with
gatings that have \emph{localized receptive fields} (i.e., the activation
of each gating unit depends on only a subset of input dimensions,
the receptive field of all gating units tile the $N_{0}$ input dimensions,
as shown in Fig.$\text{\ref{fig:Dependence-of-generalization}}$(a)
bottom), where the student GGDLN is required to learn the input-output
relation of a given ReLU teacher. For a ReLU teacher with a single
fully connected hidden layer (Fig.$\text{\ref{fig:Dependence-of-generalization}}$(a)
top left), gatings with different receptive fields are of equal importance,
hence the renormalization does not play a beneficial functional role,
and the infinitely wide network performs better than finite $N$.
As shown in Fig.$\text{\ref{fig:Dependence-of-generalization}}$(b),
bias, variance and generalization error all decrease with $N$. For
a 'local' ReLU teacher with larger input weights for one subset of
input components (the preferred inputs, Fig.$\text{\ref{fig:Dependence-of-generalization}}$(a)
top right), renormalization improves task performance by the selective
increase of the elements in ${\bf U}_{1}$ that correspond to gating
units whose receptive fields overlap the teacher's preferred inputs
(Fig.$\text{\ref{fig:Dependence-of-generalization}}$(d\&e)). Hence,
narrower networks (with a stronger renormalization) generalize better,
and both the bias and the generalization error increase with $N$
(Fig.$\text{\ref{fig:Dependence-of-generalization}}$(c)). More generally,
the input can represent a set of fixed features of the data, and the
'local' teacher generate labels depending on a subset of the features.
Therefore, networks with finite width are able to select the relevant
set of features by adjusting the amplitude in the renormalization matrix ${\bf U}_{1}$ to assign the gating units with different importance
for the task, while in the GP limit the network always assigns equal
importance to all the gating units. 

To summarize, our theory not only captures the more complex behavior
of generalization (especially bias) as a function of network width,
but also provides qualitative explanation of how generalization is
affected by the structure of the renormalization matrix in different
tasks.

\begin{figure}[h]
\centering

\includegraphics[width=0.8\textwidth]{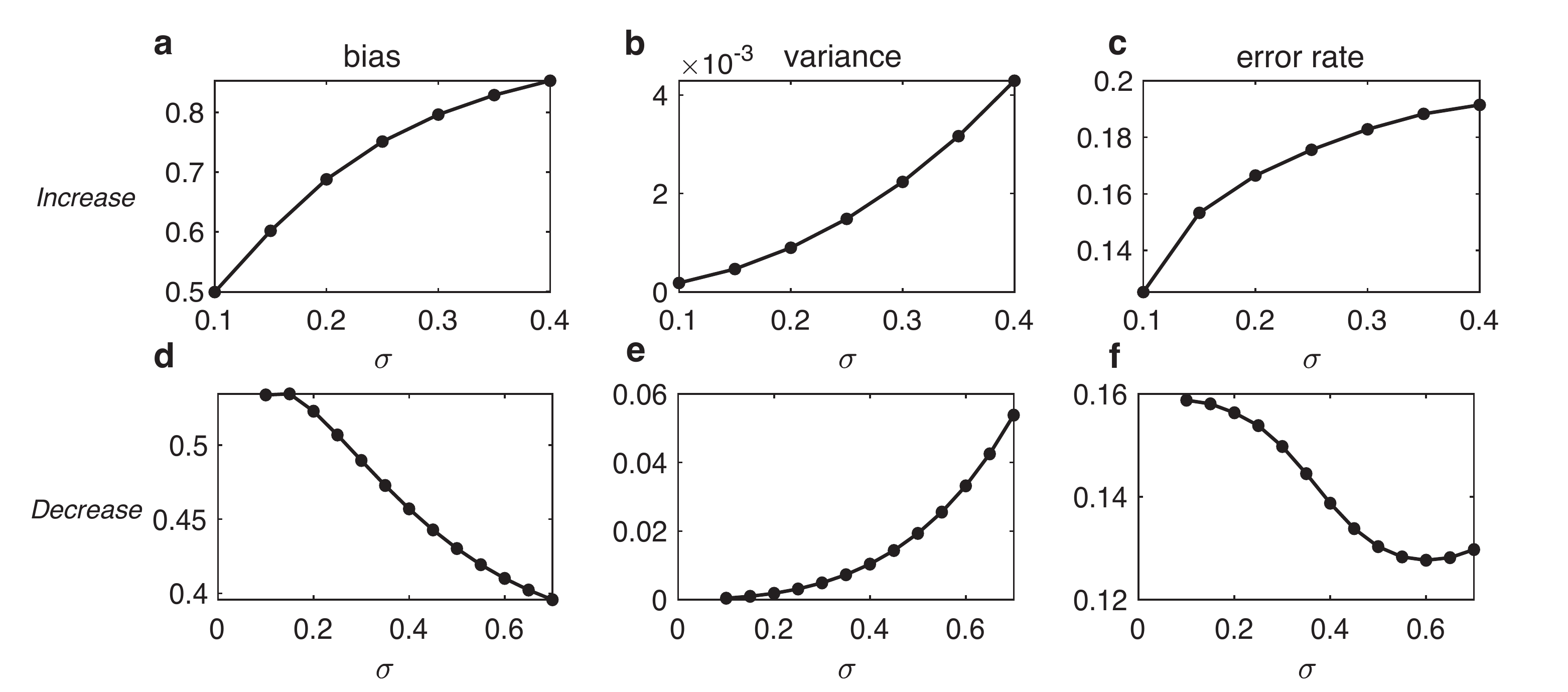}\caption{\label{fig:sigdependence}Generalization as a function of $\sigma$
for GGDLNs trained on MNIST dataset predicted by our theory. (\textbf{a}-\textbf{c})
Bias (a), variance (b) and error rate (c) increase as a function of
$\sigma$ . (\textbf{b}-\textbf{f}) Bias decreases as a function of
$\sigma$ while variance increases, leading to an optimal $\sigma$
with minimum error rate.}
\end{figure}

\textbf{Effect of regularization strengths on generalization performance:}
Similar to the dependence on $N$, generalization also exhibits different
behavior as a function of the regularization parameter $\sigma$ in
different parameter regimes, with contributions from both the bias
and the variance. The dependence of error bias on $\sigma$ also arises
due to the matrix nature of the renormalization. In Fig.$\text{\ref{fig:sigdependence} }$,
we show parameter regimes where the bias can increase (Fig.$\text{\ref{fig:sigdependence}}$
(a-c)) or decrease (Fig.$\text{\ref{fig:sigdependence}}$ (d-f)) with
$\sigma$ on MNIST dataset \cite{lecun-mnisthandwrittendigit-2010} (Appendix
$\text{\ref{subsec:MNIST-classification} }$). Although the dependence
on $\sigma$ is complicated and diverse, and there lacks a general rule
for when the qualitative behavior changes, we found that our theory
accurately captures the qualitative behavior of results obtained from
GD (Appendix $\text{\ref{subsec:Gradient-descent-dynamics}}$ Fig.$\text{\ref{fig:Simulation-with-gd1}}$).
In both regimes the variance increases with $\sigma$ as the solution
space expands for a weaker regularization. Specifically in Fig.$\text{\ref{fig:sigdependence} (d-f)}$,
due to the increasing variance (e) and decreasing bias (d), there
is a minimum error rate ((f), Appendix $\text{\ref{subsec:Generalization}}$
Eq.$\text{\ref{eq:errorrate}}$ for how error rate is calculated from
the mean and variance of the predictor) at intermediate $\sigma$,
indicating an optimal level of regularization strength as opposed
to linear networks \cite{li2021statistical}, where strong regularization
($\sigma=0$) always results in optimal generalization

\begin{figure}[h]
\centering
\includegraphics[width=0.7\textwidth]{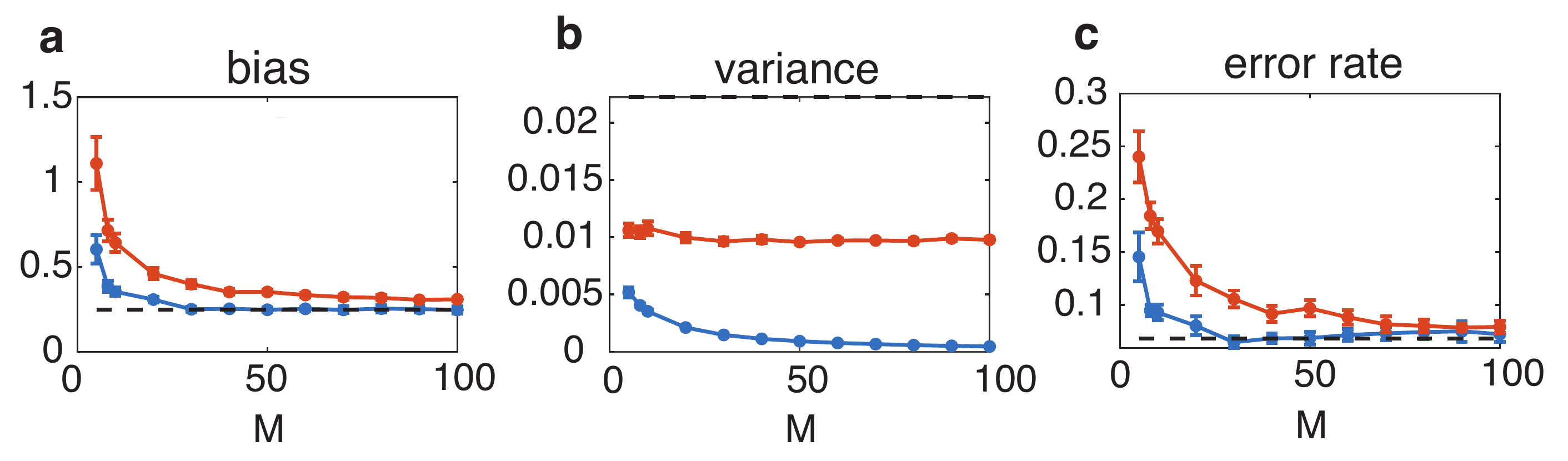}

\caption{\label{fig:pretrainedgatings}Dependence of generalization on M for
GGDLNs trained on MNIST dataset predicted by our theory. Bias (a),
variance (b) and error rate (c) as a function of M for random (red
lines) and pretrained gatings (blue lines), and ReLU network with
the same width (black dashed lines).}
\end{figure}

\textbf{GGDLNs with different choices of gatings achieve comparable
performance to ReLU networks:} The nonlinear operation of the gatings
enables the network to learn nonlinear tasks. In Fig.$\text{\ref{fig:pretrainedgatings}}$,
we show that although the gatings are fixed during training, the network
achieves comparable performance as a fully trained nonlinear (ReLU)
network with the same hidden layer width for classifying even and
odd digits in MNIST data when $M$ is sufficiently large (over-parameterization does not lead to over-fitting here, as shown also in other nonlinear networks \cite{belkin2019reconciling,belkin2020two}, possibly due to the explicit $L_2$ prior). Furthermore, although the gatings are fixed during the supervised
training of the GGDLN, they can be cleverly chosen to improve generalization
performance. To demonstrate this strategy, we compared two different
choices of gatings. \emph{Random gatings} take the form $g_{m}({\bf x})=\Theta(\frac{1}{\sqrt{N_{0}}}{\bf V}_{m}^{\top}{\bf x}-b)$,
where ${\bf V}_{m}$ is a $N_{0}$-dimensional random vector with standard
Gaussian i.i.d. elements, $b$ is a scalar threshold, and $\Theta(x)$
is the heaviside step function. The \emph{pretrained gatings} are
trained on the \emph{unlabelled} training dataset with unsupervised
soft k-means clustering, such that the $m$-th gating $g_{m}({\bf x})$
outputs the probability of assigning data ${\bf x}$ to the $m$-th cluster
(Appendix $\text{\ref{subsec:MNIST-classification}}$). As shown in
Fig.$\text{\ref{fig:pretrainedgatings}}$, for pretrained gatings,
generalization performance improves with $M$ much faster compared
to random gatings, and approaches the performance of ReLU network
at a smaller $M$. Our theory (Fig.$\text{\ref{fig:pretrainedgatings}}$)
and numerical results of GD dynamics (Appendix $\text{\ref{subsec:Gradient-descent-dynamics}}$
Fig.$\text{\ref{fig:Simulation-with-gd2}}$) agree qualitatively well.
The result shows that GGDLNs can still achieve competitive performance
on nonlinear tasks while remaining theoretically amenable.

\subsection{Kernel shape renormalization in deeper networks}

We now consider the effect of the matrix renormalization on GGDLNs
with more layers. We begin by analyzing the renormalization effect on the \textit{shape} of the kernel in deep architectures. It is well known
that the GP kernel of many nonlinear networks flattens (the kernel function goes to a constant) as network
depth increases \cite{lee2017deep}, ultimately losing information
about the input and degrading generalization performance. Here we
show that kernel shape renormalization slows down flattening of kernels
by incorporating data relevant information into the learned weights.

To study the \emph{shape} of the kernel independent of kernel magnitude,
we define the normalized kernel $\mathcal{K}_{L}({\bf x},{\bf y})=\frac{\tilde{K}_{L}({\bf x},{\bf y})}{\tilde{K}_{L}({\bf x},{\bf x})^{1/2}\tilde{K}_{L}({\bf y},{\bf y})^{1/2}}$,
where $\tilde{K}_{L}({\bf x},{\bf y})$ denotes the renormalized kernel
for GGDLN with $L$ hidden layers. This normalized kernel measures
the cosine of the vectors ${\bf x}$ and ${\bf y}$ with generalized
inner product defined by the kernel $\tilde{K}_{L}({\bf x},{\bf y})$,
and therefore $\mathcal{K}_{L}({\bf x},{\bf y})\in[-1,1]$. For the GP
kernel of GGDLNs, we have $\mathcal{K}_{L}({\bf x},{\bf y})=\cos({\bf g}({\bf x}),{\bf g}({\bf y}))^{L}\cos({\bf x},{\bf y})$.
While $\mathcal{K}_{L}$ depends on the specific choice of gatings in
general, in the special case of \textit{random gatings} with zero
threshold $g_{m}({\bf x})=\Theta(\frac{1}{\sqrt{N_{0}}}{\bf V}_{m}^{\top}{\bf x})$
and the number of gatings $M\rightarrow\infty$, we can write $\mathcal{K}_L$
analytically as a function of the angle $\theta$ between input vectors
$x$ and $y$, given by $\mathcal{K}_{L}(\theta)=(\frac{\pi-\theta}{\pi})^{L}\cos(\theta)$, $\theta\in[-\pi,\pi]$.
Thus, as $L\rightarrow\infty$, $\mathcal{K}_{L}(\theta)$ shrinks to zero except for $\theta=0$. This
'flattening' effect reflects the loss of information in deep networks, as pairs of inputs with different similarities now all have hidden representations that are orthogonal. The effect also empirically holds true for networks with
finite $M$ (see Appendix $\text{\ref{subsec:Deep-GP-kernel}}$).

\begin{figure}[h]
\centering

\includegraphics[width=1\textwidth]{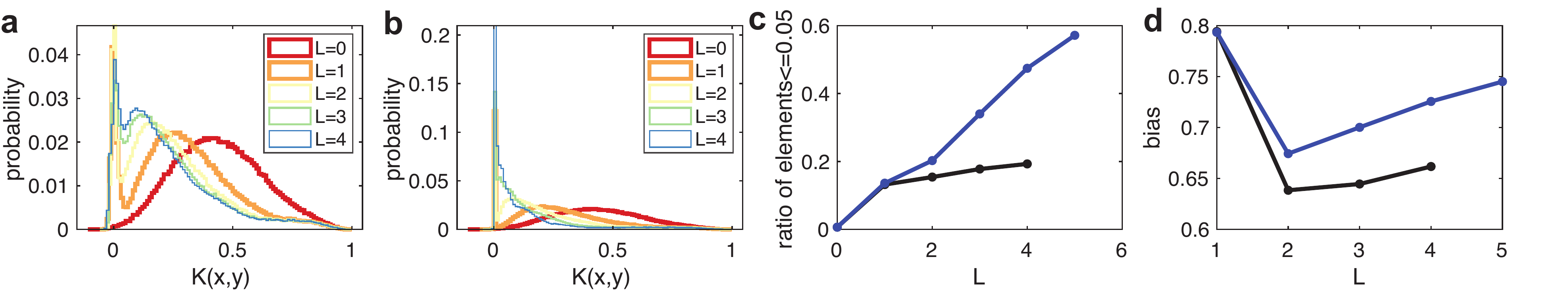}\caption{\label{fig:Shape-renormalization-slows}Shape renormalization slows
down flattening of kernels in deep networks. (\textbf{a}-\textbf{b})
Distribution of kernel elements $\mathcal{K}_{L}({\bf x},{\bf y})$ for
the renormalized kernel (a) and GP kernel (b) for different network
depth $L$. (\textbf{c}) Ratio of kernel elements smaller or equal
to 0.05 increases faster for GP kernel (blue line) compared to the
renormalized kernel (black line), the renormalization slows down the
rate at which elements in the GP kernel shrink to zero as a function
of $L$. (\textbf{d}) The bias contribution to the generalization
first decreases then increases as a function of $L$ due to the flattening
of the kernel (blue line). Finite width network with renormalized
kernel performs better for $L>1$ in this parameter regime (black
line). See Appendix $\text{\ref{subsec:MNIST-classification}}$ for
detailed parameters.}
\end{figure}

In Fig.$\text{\ref{fig:Shape-renormalization-slows} },$we study the
effect of kernel renormalization on the 'flattening' effect of deep GGDLNs.
As shown in Fig.$\text{\ref{fig:Shape-renormalization-slows} }$(a)-(c),
the elements of the renormalized kernel shrink to zero at a much slower
rate compared to the GP kernel. (Note that unlike the variance, the
bias is affected only by shape changes, but not by changes in the
amplitude of the kernel, in Fig.$\text{\ref{fig:Shape-renormalization-slows}}$(d)
we plot only the bias contribution to the generalization.) While mitigating
the flattening of the GP kernel is a general feature of our renormalized
kernel for different parameters, its effect on the generalization
performance (especially the bias) may be different for different network
parameters. In the specific example in Fig.$\text{\ref{fig:Shape-renormalization-slows}}$,
finite width networks with a less 'flattened' renormalized kernel
achieve better performance than the GP limit. Both the GP limit and
the finite width networks have optimal performance at $L=2$ in this
example. 

\vspace{-0.2cm}
\section{GGDLNs for multiple tasks}

In this section, we apply our theory to investigate the ability of
GGDLNs to perform multiple tasks. We consider two different scenarios
below. First, different tasks require the network to learn input-output
mappings on input data with different statistics. This scenario corresponds
to real life situations where the training data distribution is non-stationary.
The tasks can be separated without any additional top-down information. In this case, the gatings
are bottom-up, and are functions of the input data only. In the second
case, different tasks give conflicting labels for the same inputs,
corresponding to the situation where performing the two tasks require
additional top-down contextual information, and the information can
be incorporated into the gating units in GGDLNs. In both scenarios,
when the gatings are fixed and we modulate the de-correlation by changing
network width and thus the strength of the kernel renormalization,
we find that de-correlation between tasks leads to better
generalization performance.

\subsection{Bottom-up gating units}

\begin{SCfigure}[][h]

\includegraphics[width=0.58\textwidth]{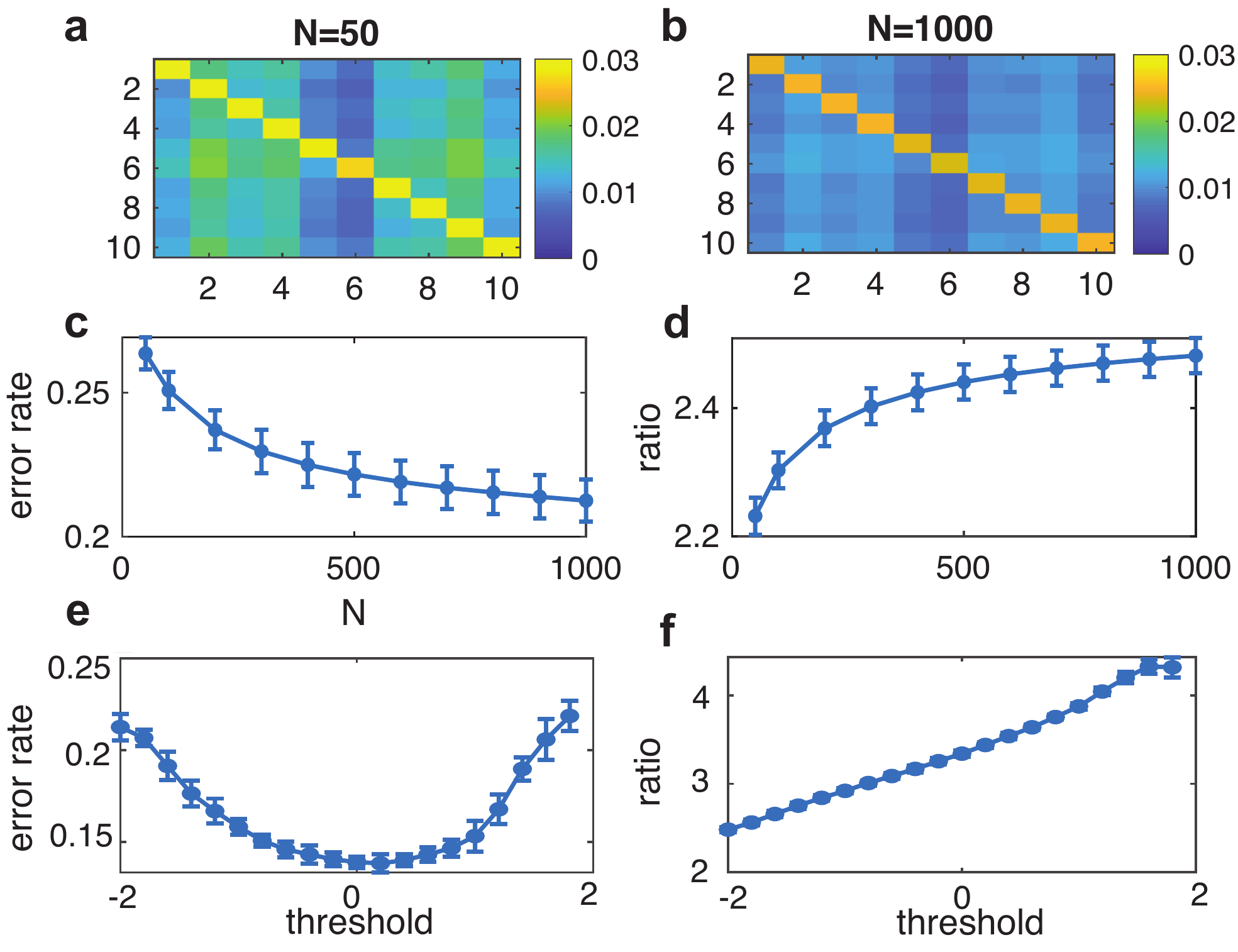}\caption{GGDLNs with bottom-up gating units learning multiple tasks trained on permuted MNIST. (\textbf{a}-\textbf{b}) Task-task correlation matrix C for $N=50$ and $N=1000$, different permutations are more decorrelated for larger N. (\textbf{c}) Error rate decreases as a function of N due to the decorrelation. (\textbf{d}) Ratio of the average amplitude of diagonal elements versus off-diagonal elements in C increases as a function of N. (\textbf{e}) Error rate first decreases then increases as a function of gating threshold. (\textbf{f}) Decorrelation increases as a function of gating threshold.}
\label{fig:permutedmnist}
\end{SCfigure}First we consider learning different tasks defined by vastly different
input statistics with bottom-up gatings, using permuted MNIST as an
example. Previous works have shown that GLNs mitigate catastrophic
forgetting when sequentially trained on permuted MNIST \cite{sezener2021rapid}.
While our theory does not address directly the dynamics of sequential
learning, we aim to shed light on this question by asking
how the two tasks interfere with each other when they are learned
simultaneously.

We introduce a measure of \emph{inter-task interference }by noting
that after learning the mean predictor on a new data ${\bf x}$ ,
Eq.$\text{\ref{eq:predictorstatistics}}$, is a linear combination
of the output labels $Y^{\mu}$ of all the training data, and the
coefficient of this linear combination, is given by the $\mu$-th
coefficient of $\tilde{{\bf k}}({\bf x})^{\top}\tilde{{\bf K}}^{-1}$.
Thus, we define a task-task correlation matrix, via $C_{pq}=\sum_{\mu=1}^{P}\sum_{\gamma=1}^{P_{t}}\rvert\tilde{{\bf k}}^{T}\tilde{{\bf K}}^{-1}\lvert_{p\gamma,q\mu}$($p,q=1,\cdots,n$),
where we assume there are $P$ training examples and $P_{t}$ test
data for each task, with a total of $n$ tasks. The amplitude of each
element $C_{pq}$ measures how much \emph{training data of task $q$}
contribute to the prediction on the \emph{test data of task $p$}.
Stronger diagonal elements indicates that the network separates the
processing of data of different tasks (Fig.$\text{\ref{fig:permutedmnist}}$(a)-(b)).
As we show in Fig.$\text{\ref{fig:permutedmnist}}$, we can tune the
relative strength of the diagonal elements of ${\bf C}$ smoothly
by changing the network width (Fig.$\text{\ref{fig:permutedmnist}}$(a)-(d))
or by changing the threshold of the gating (Fig.$\text{\ref{fig:permutedmnist} }$(e)-(f)).
In the case where the gatings are fixed and the network width is changed,
an increase in the strength of the diagonal elements (Fig.$\text{\ref{fig:permutedmnist}}$(d))
results in better generalization (Fig.$\text{\ref{fig:permutedmnist}}$(c)),
indicating that the network generalizes better by processing data
of different tasks separately through the gating units. However,
in the case where we change the activation of the gatings by adjusting
the threshold, although different tasks are more de-correlated when
the threshold is large due to a set of less overlapping gatings activated
for each task, generalization error first decreases and then increases
again. This is because for large threshold the sparsity of the gatings
activated for each task limits the nonlinearity of the network, and
therefore the generalization performance on this nonlinear task. 

\subsection{\label{subsec:Combined-top-down-and}Combined top-down and bottom-up
gating units}

\begin{figure}[h]
\centering

\includegraphics[width=0.9\textwidth]{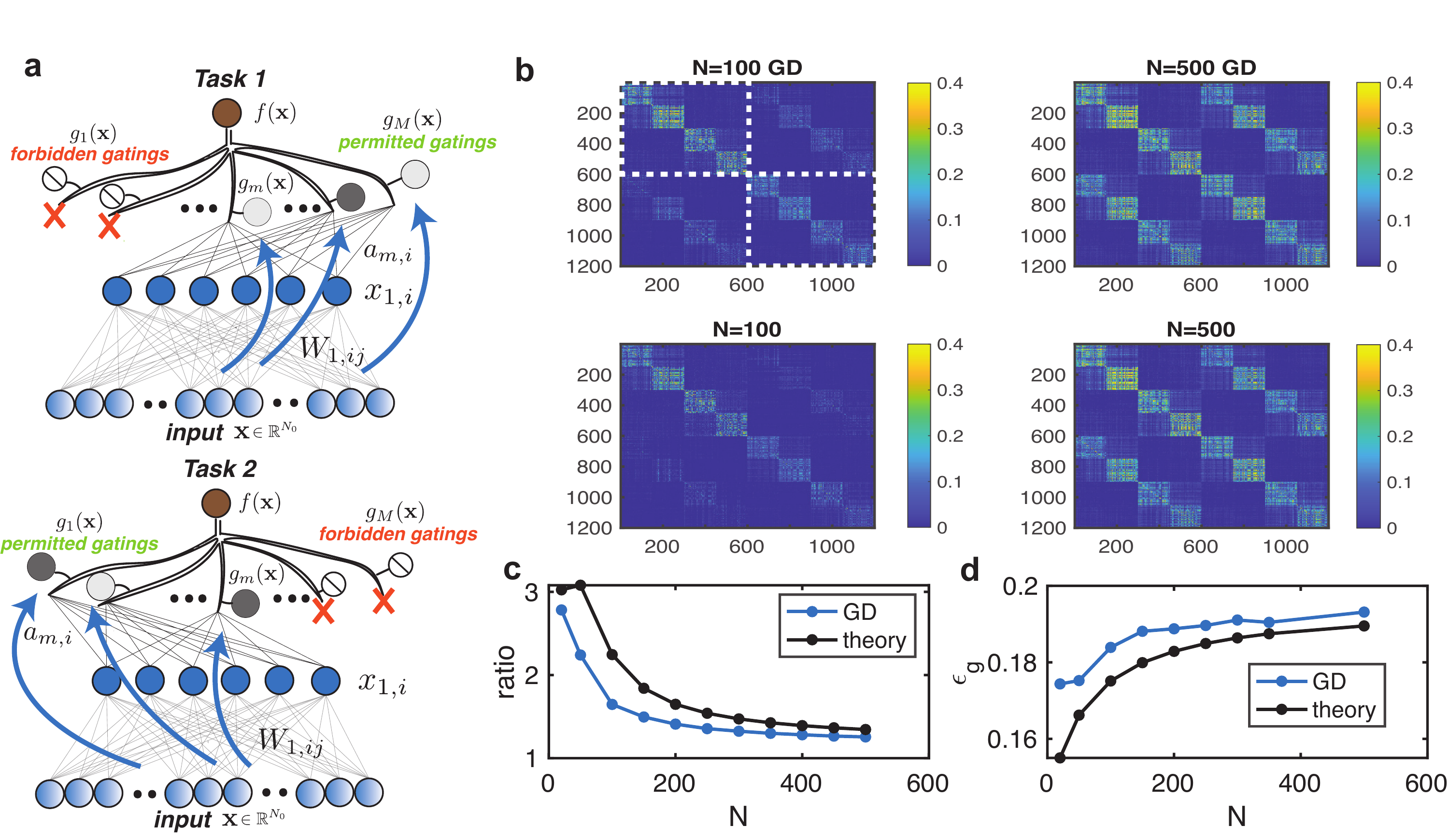}

\caption{\label{fig:Kernel-renormalization-decorrela}Kernel renormalization
de-correlates different tasks defined by different labels on the same
inputs. (\textbf{a}) GGDLNs performing two tasks using combined top-down
and bottom-up task signal.  (\textbf{b}) Top: Renormalized kernel
calculated with Eq. $\text{\ref{eq:interpretation}}$ from GD dynamics.
Bottom: Renormalized kernel theory. (\textbf{c}) Ratio of the magnitude
of diagonal (blocks with white dashed lines in (b)) versus off diagonal blocks
decreases as a function of $N$. (\textbf{d}) Generalization error increases
with N.}
\end{figure}

We now consider learning two tasks that provide conflicting labels
on the same input data. The gating units combine both top-down task signal which informs the system of which task to perform
for a given input, and bottom-up signals which, as before, depend
on the input. In different tasks, different sets of gatings are \emph{permitted
or forbidden} depending on the top-down signal, then the states of
the permitted gatings are further determined as a function of the
input ${\bf x}$, while the forbidden gatings are set to $0$, and the corresponding dendritic branches do not connect to the previous
layer neurons (Fig.$\text{\ref{fig:Kernel-renormalization-decorrela}}$(a))
in this task. For a single hidden layer network, with a similar argument
as in Section $\text{\ref{sec:Globally-gated-deep}}$, it is straightforward
to show that the number of different tasks that can be memorized is
given by $n\leq M$ and the number of training examples for each task
needs to satisfy $P\leq N_{0}M_{p}$, where $M_{p}$ is the number
of permitted gating units in each task. In the limiting case where
a set of non-overlapping gating units are permitted in each of the
$n$ tasks, the network is equivalent to $n$ sub-networks, each independently
performing one task. In this case $M_{p}$ is limited by $M/n$, which
in turn limits the capacity and the effective input-output nonlinearity
for each independent task. We consider the case where the permitted
gatings are chosen randomly for each task and are therefore in general
overlapping across tasks. We then investigate how learning modifies the correlation induced
by the overlapping gatings through the renormalization matrix. As an example we consider training on
permuted and un-permuted MNIST digits of $0$ and $1$'s. One task
is to classify the two digits in both permuted
and un-permuted data, and the second task is to separate the permuted
digits (both 0 and 1) from the un-permuted digits. The labels of the
two tasks are uncorrelated, while the permitted gatings of the two tasks are partially overlapping. In this case the renormalized
kernel $\tilde{{\bf K}}_{1}$ can be written as $\tilde{K}_{p\mu,q\nu}=(\frac{1}{M}{\bf g}^{p}({\bf x}^{\mu})^{\top}{\bf U}_{1}{\bf g}^{q}({\bf x}^{\nu}))\frac{\sigma^{2}}{N_{0}}{\bf x}^{\mu\top}{\bf x}^{\nu}$.
Here $p,q\in\{1,2\}$ are the task indices, and $\mu$,$\nu=1,\cdots,P$
are the input indices. The kernel is therefore $2P\times2P$ as shown
in Fig.$\text{\ref{fig:Kernel-renormalization-decorrela}}$(b) ($P=600$); the
diagonal blocks (white dashed lines) correspond to kernels of task 1 and task 2,
while the off diagonal blocks correspond to the cross kernels. In Fig.$\text{\ref{fig:Kernel-renormalization-decorrela}}$(b) bottom,
we show the renormalized kernel with the renormalization matrix ${\bf U}_{1}$
calculated by solving Eq.$\text{\ref{eq:spequation}}$. Similar results
are achieved by by numerically estimating Eq.$\text{\ref{eq:interpretation} }$ with readout weights obtained from GD dynamics (Fig.$\text{\ref{fig:Kernel-renormalization-decorrela}}$(b) top).

The results demonstrate that stronger kernel renormalization achieved
in narrower networks suppresses more strongly the correlation between
tasks, reflected by the weaker off-diagonal blocks in Fig.$\text{\ref{fig:Kernel-renormalization-decorrela}}$(b). A decreasing ratio between the average amplitudes of the diagonal
and off-diagonal blocks shows that the de-correlation effect diminishes
for large $N$, leading to increasing generalization error with $N$(Fig.$\text{\ref{fig:Kernel-renormalization-decorrela}}$(c
\& d)).
\vspace{-0.2cm}
\section{Discussion}\label{sec:Discussion}
In this work, we proposed a novel gating network architecture, the
GGDLN, amenable to theoretical analysis of the network expressivity
and generalization performance. The predictor statistics of GGDLNs
can be expressed in terms of kernels that undergo shape renormalization,
resulting diverse behavior of the bias as a function of various network
parameters. This renormalization slows down the flattening of the
GP kernel in deep networks, suggesting that the loss of input information
as $L$ increases may be prevented in finite-width nonlinear networks.
We also investigate the capability of GGDLNs to perform multiple tasks.
While our theory is an exact description of the posterior of weight
distribution induced by Langevin dynamics in Bayesian learning, it
provides surprisingly well qualitative agreement with results obtained
with GD dynamics for not only the generalization but also the kernel
representation with matrix renormalization,
largely extending its applicability. There are several limitations of our work. Our mean-field analysis
is accurate in the ‘finite-width’ thermodynamic limit where both $P$
and $N$ go to infinity, but $M$ and $L$ remain finite. In practice,
the size of the renormalization matrix increases as $M^L$, hence for some moderate $M$, as $L$ increases, any large but finite $N$
might eventually get the network outside the above thermodynamic regime.
The theory also focuses on the equilibrium distribution induced by
learning and does not address important questions related to the learning
dynamics. Finally, although we have shown qualitative correspondence
of the GGDLN properties and standard DNNs with local nonlinearity,
as ReLU, a full theory of the thermodynamic limit of DNNs with local
nonlinearity is still an open challenge.

While our theory currently addresses learning in GGDLNs using a \textit{global}
cost function, exploring the possibility of extending the formalization
of the equilibrium distribution to characterize local learning dynamics is an ongoing work. Recent works have shown
that multilayer perceptrons (MLPs) with learned gatings that implements
spatial attention have surprisingly good performance on Natural Language
Processing (NLP) and computer vision \cite{liu2021pay}. Extension
of our theory to learnable gatings that implements attention mechanisms
remains to be explored. Furthermore, incorporating convolutional architecture \cite{van2017convolutional,naveh2021self,gu2018recent,garriga2018deep,albawi2017understanding} into our GGDLNs and using
the gating units to encode context-dependent modification of different
feature maps is an interesting direction related to the fast-developing
research topic of visual question-answering (VQA) \cite{antol2015vqa,kafle2017visual,lu2016hierarchical}
, where answering different questions about the same image is similar
to performing multiple tasks in different contexts with different
labels on the same dataset, as we discussed in Section $\text{\ref{subsec:Combined-top-down-and}}$.
We leave these exciting research directions for future work.
\vspace{-0.2cm}
\section*{Acknowledgement}
We thank the anonymous reviewers for their helpful comments. This research is supported by the Swartz Foundation, the NIH grant from the NINDS (No. 1U19NS104653), and the Gatsby Charitable Foundation. We acknowledge the support of a generous gift from Amazon. This paper is dedicated to the memory of Mrs. Lily Safra, a great supporter of brain research.

\bibliographystyle{unsrt}
\bibliography{reference}

\section*{Checklist}
\begin{enumerate}
\item For all authors... 
\begin{enumerate}   
\item Do the main claims made in the abstract and introduction accurately reflect the paper's contributions and scope? \answerYes{}   
\item Did you describe the limitations of your work? \answerYes{See Section \ref{sec:Kernel-shape-renormalization} and \ref{sec:Discussion}}   
\item Did you discuss any potential negative societal impacts of your work? \answerNA{}   
\item Have you read the ethics review guidelines and ensured that your paper conforms to them?  \answerYes{} 
\end{enumerate}
\item If you are including theoretical results... 
\begin{enumerate}   
\item Did you state the full set of assumptions of all theoretical results?  \answerYes{See Section \ref{sec:Kernel-shape-renormalization}}         
\item Did you include complete proofs of all theoretical results?     \answerYes{See Appendix \ref{sec:Derivation}} 
\end{enumerate}
\item If you ran experiments... 
\begin{enumerate}   
\item Did you include the code, data, and instructions needed to reproduce the main experimental results (either in the supplemental material or as a URL)?   \answerYes{In supplementary material}   
\item Did you specify all the training details (e.g., data splits, hyperparameters, how they were chosen)? \answerYes{See Appendix \ref{sec:Detailed-parameters}}         
\item Did you report error bars (e.g., with respect to the random seed after running experiments multiple times)? \answerYes{See Fig.\ref{fig:permutedmnist}}         
\item Did you include the total amount of compute and the type of resources used (e.g., type of GPUs, internal cluster, or cloud provider)?   \answerYes{See Appendix \ref{sec:Detailed-parameters}} 
\end{enumerate}
\item If you are using existing assets (e.g., code, data, models) or curating/releasing new assets... 
\begin{enumerate}   
\item If your work uses existing assets, did you cite the creators?   \answerYes{}   
\item Did you mention the license of the assets?     \answerYes{Appendix \ref{sec:Detailed-parameters}}   
\item Did you include any new assets either in the supplemental material or as a URL?     \answerNo{}   
\item Did you discuss whether and how consent was obtained from people whose data you're using/curating?     \answerNA{}   
\item Did you discuss whether the data you are using/curating contains personally identifiable information or offensive content?     \answerNA{} 
\end{enumerate}
\item If you used crowdsourcing or conducted research with human subjects... 
\begin{enumerate}   
\item Did you include the full text of instructions given to participants and screenshots, if applicable?     \answerNA{}   
\item Did you describe any potential participant risks, with links to Institutional Review Board (IRB) approvals, if applicable?     \answerNA{}   
\item Did you include the estimated hourly wage paid to participants and the total amount spent on participant compensation?     \answerNA{} 
\end{enumerate}
\end{enumerate}

\newpage{}

\appendix

\setcounter{page}{1} 
\setcounter{section}{0} 
\setcounter{equation}{0} 
\setcounter{figure}{0}

\begin{center} 	
\begin{Large}      
\bf{Supplementary Information (SI)}      
\end{Large} 
\end{center}

\section{\label{sec:Derivation}Derivation of the kernel shape renormalization
theory}

\subsection{\label{subsec:Kernel-shape-renormalization}Kernel shape renormalization}

\paragraph{Derivation of the theory.}

We begin with the partition function 
\begin{equation}
Z=\int d\Theta\exp[-\frac{1}{2T}\sum_{\mu=1}^{P}(\frac{1}{\sqrt{NM}}\sum_{i=1}^{N}\sum_{m=1}^{M}a_{m,i}x_{L,i}^{\mu}g_{m}({\bf x}^{\mu})-Y^{\mu})^{2}-\frac{1}{2\sigma^{2}}{\bf \Theta}^{\top}{\bf \Theta}]
\end{equation}
and introduce $P$ auxilliary integration variables, $t^{\mu}(\mu=1,\cdots,P)$,
to linearize the quadratic training error. 
\begin{equation}
Z=\int d\Theta\int\Pi_{\mu=1}^{P}dt_{\mu}\exp[-\frac{1}{2\sigma^{2}}{\bf \Theta}^{\top}{\bf \Theta}-\sum_{\mu=1}^{P}it_{\mu}(\frac{1}{\sqrt{N}}\sum_{i=1}^{N}a_{m,i}x_{L,i}^{\mu}g_{m}({\bf x}^{\mu})-Y^{\mu})^{2}-\frac{T}{2}{\bf t}^{\top}{\bf t}]
\end{equation}
Integrate over ${\bf a}$ in the $T\rightarrow0$ limit, we have 
\begin{equation}
Z=\int d{\bf W}\text{\ensuremath{\int\Pi_{\mu=1}^{P}dt_{\mu}}\ensuremath{\exp[-\frac{1}{2}{\bf t}^{\top}{\bf K}_{L}({\bf W}){\bf t}+i{\bf t}^{\top}{\bf Y}-\frac{1}{2\sigma^{2}}\mathrm{Tr}({\bf W}^{\top}{\bf W})]}}
\end{equation}
with $K_{L}^{\mu\nu}({\bf W})=(\frac{\sigma^{2}}{M}{\bf g}({\bf x}^{\mu})^{\top}{\bf g}({\bf x}^{\nu}))(\frac{1}{N}{\bf x}_{L}^{\mu}({\bf W})^{\top}x_{L}^{\nu}({\bf W}))$.
Integrating over $t$ yields
\begin{equation}
Z=\int d{\bf W}\exp[-\frac{1}{2\sigma^{2}}\mathrm{Tr}({\bf W}^{\top}{\bf W})+\frac{1}{2}{\bf Y}^{\top}{\bf K}_{L}({\bf W})^{-1}{\bf Y}+\frac{1}{2}\log\det({\bf K}_{L}({\bf W}))]\label{eq:z(W)}
\end{equation}
In the single hidden layer case, ${\bf x}_{L}^{\mu}={\bf x}_{1}^{\mu}=\frac{1}{\sqrt{N_{0}}}{\bf W}_{1}^{\top}{\bf x}^{\mu}$.
We can integrate out ${\bf W}_{1}$(the first hidden layer weights)
in the thermodynamic limit due to this linear relation, and obtain
\begin{equation}
Z=\int\Pi_{\mu=1}^{P}dt_{\mu}\exp[i{\bf t}^{\top}{\bf Y}+NG({\bf t})],\ G({\bf t})=\log\langle\exp-\frac{1}{2N}{\bf t}^{\top}{\bf K}_{L}^{{\bf w}}{\bf t}\rangle_{{\bf w}}\label{eq:g(t)}
\end{equation}
where the average is with respect to a \textit{single} $N_{0}$-dimensional
weight vector ${\bf w}_{1,i}$ with i.i.d. $\mathcal{N}(0,\sigma^{2})$
components, and $K_{L}^{{\bf w},\mu\nu}=K_{1}^{{\bf w},\mu\nu}=\frac{\sigma^{2}}{M}{\bf g}({\bf x}^{\mu})^{\top}{\bf g}({\bf x}^{\nu})\frac{1}{N_{0}}{\bf x}^{\mu\top}{\bf w}_{1,i}{\bf w}_{1,i}^{\top}{\bf x}^{\nu}$.
The term in the exponent of SI Eq. $\text{\ref{eq:g(t)} }$ is quadratic
in ${\bf w}_{L,i}$ and therefore
\begin{align}
\langle\exp-\frac{1}{2N}{\bf t}^{\top}{\bf K}_{L}^{{\bf w}}{\bf t}\rangle_{{\bf w}} & =\int d{\bf w}_{1,i}\exp(-\frac{1}{2}{\bf w}_{1,i}^{\top}{\bf M}{\bf w}_{1,i})\\
M_{jk} & =\sigma^{-2}\delta_{ij}+\frac{1}{N}\sum_{\mu\nu}t_{\mu}t_{\nu}\frac{\sigma^{2}}{M}{\bf g}({\bf x}^{\mu})^{\top}{\bf g}({\bf x}^{\nu})\frac{1}{N_{0}}x_{j}^{\mu}x_{k}^{\nu}\label{eq:M}
\end{align}
performing this averaging yields $G({\bf t})=-\frac{1}{2}\log\det({\bf I}+{\bf \mathcal{H}}_{1})$
\begin{equation}
{\bf \mathcal{H}}_{1}^{mn}=\frac{1}{N}\sum_{\mu\nu}t_{\mu}t_{\nu}\frac{\sigma^{2}}{M}g_{m}({\bf x}^{\mu})g_{n}({\bf x}^{\nu})K_{0}^{\mu\nu}\label{eq:hmn}
\end{equation}
where $K_0^{\mu\nu}=\frac{\sigma^2}{N_0}{\bf x}^{\mu\top}{\bf x}^\nu$. To integrate over $t$, we enforce the identity SI Eq.$\text{\ref{eq:hmn}}$,
by Fourier representation of the $\delta$-function, introducing the
auxiliary variable ${\bf U}_{1}$. 
\begin{equation}
Z=\int d{\bf U}_{1}\int d\mathcal{H}_{1}\int d{\bf t}\exp[i{\bf t}^{\top}{\bf Y}-\frac{N}{2}\log\det({\bf I}+\mathcal{H}_{1})+\frac{N}{2\sigma^{2}}\mathrm{Tr}({\bf U}_{1}\mathcal{H}_{1})-\frac{1}{2}{\bf t}^{\top}\tilde{{\bf K}}_{1}{\bf t}]
\end{equation}
with $\tilde{K}_{1}^{\mu\nu}=(\frac{1}{M}{\bf g}({\bf x}^{\mu})^{\top}{\bf U}_{1}{\bf g}({\bf x}^{\nu}))K_{0}^{\mu\nu}$
as defined in main text. Integrate over $t$, 
\begin{align}
Z & =\int d{\bf U}_{1}\int d\mathcal{H}_{1}\exp[-\frac{N}{2}\log\det({\bf I}+\mathcal{H}_{1})+\frac{N}{2\sigma^{2}}\mathrm{Tr}({\bf U}_{1}\mathcal{H}_{1})-\frac{1}{2}{\bf Y}^{\top}\tilde{{\bf K}}_{1}^{-1}{\bf Y}-\frac{1}{2}\log\det\tilde{{\bf K}}_{1}]\label{eq:partition-integrated}
\end{align}
In the limit of $N\rightarrow\infty$,$P\rightarrow\infty$, and finite
$\alpha=P/N$, we can solve this integral with the saddle-point method.
One of the saddle-point equations (by taking derivative w.r.t. $\mathcal{H}_{1}$)
yields ${\bf U}_{1}=\sigma^{2}(I+\mathcal{H}_{1})^{-1}$. Plugging
back into SI Eq.$\text{\ref{eq:partition-integrated} }$, we obtain
$Z=\int d{\bf U}_{1}\exp(-H_{1})$, with the effective Hamiltonian
given by Eq.$\text{\ref{eq:hamiltonian} }$ in the main text, integration
over ${\bf W}_{1}$ results in the presence of an auxiliary ${\bf U}_{1}$,
which can be eliminated in the thermodynamic limit through the saddle-point
equation Eq.$\text{\ref{eq:spequation}}$. 

\paragraph{Interpretation of the order parameter ${\bf U}_{1}$. }

The auxiliary integration variable ${\bf U}_{1}$ has a simple physical
interpretation. We prove Eq.$\text{\ref{eq:interpretation} }$ by
calculating $\langle\frac{1}{N}\sum_{i}a_{m,i}a_{n,i}\rangle$. 
\begin{align}
 & \langle\frac{1}{N}\sum_{i}a_{m,i}a_{n,i}\rangle=\int d\Theta\int\Pi_{\mu}^{P}dt_{\mu}\frac{1}{N}\sum_{i}a_{m,i}a_{n,i}\\
 & \exp[-\frac{1}{2\sigma^{2}}\Theta^{\top}\Theta-\sum_{\mu=1}^{P}it_{\mu}(\frac{1}{\sqrt{NM}}\sum_{i=1}^{N}\sum_{m=1}^{M}a_{m,i}x_{1,i}^{\mu}g_{m}({\bf x}^{\mu})-Y^{\mu})]\\
 & =\sigma^{2}\delta_{mn}-\frac{1}{N}\int d{\bf W}_{1}\int\Pi_{\mu=1}^{P}dt_{\mu}\sum_{jj'}\sum_{i}U_{mj}U_{nj'}W_{1,ij}W_{1,ij'}\exp-\frac{1}{2N}{\bf t}^{T}{\bf K}{\bf t}+i{\bf t}^{T}{\bf Y}
\end{align}
where
\begin{equation}
U_{mj}=\sqrt{\frac{\sigma^{4}}{NMN_{0}}}\sum_{\mu}g_{m}({\bf x}^{\mu})x_{j}^{\mu}t^{\mu}
\end{equation}
so we can write the above equation as
\begin{align}
\langle\frac{1}{N}\sum_{i}a_{m,i}a_{n,i}\rangle & =\sigma^{2}\delta_{mn}-\langle\sum_{jj'}U_{mj}U_{nj'}\langle{\bf w}{\bf w}^{\top}\rangle_{{\bf w}}^{j'j}\rangle_{{\bf t}}
\end{align}
where $w$ denotes a \textit{single} $N_{0}$-dimensional weight vector,
the average over $w$ is given by 
\begin{equation}
\langle{\bf w}{\bf w}^{\top}\rangle_{{\bf w}}=\int d{\bf w}{\bf w}{\bf w}^{\top}\exp-\frac{1}{2N}{\bf t}^{\top}{\bf K}_{1}^{{\bf w}}{\bf t}-\frac{1}{2\sigma^{2}}{\bf w}^{\top}{\bf w}+i{\bf t}^{\top}{\bf Y}
\end{equation}
similarly as in SI Eq.$\text{\ref{eq:M}}$
\begin{equation}
\frac{1}{N}{\bf t}^{\top}{\bf K}_{1}^{{\bf w}}{\bf t}+\frac{1}{\sigma^{2}}{\bf w}^{\top}{\bf w}=\sum_{jj'}w_{j}M_{jj'}w_{j'}
\end{equation}
\begin{equation}
M_{jj'}=\sigma^{-2}\delta_{jj'}+\frac{1}{N}\sum_{\mu\nu}t_{\mu}t_{\nu}\frac{\sigma^{2}}{M}{\bf g}({\bf x}^{\mu})^{\top}{\bf g}({\bf x}^{\nu})\frac{1}{N_{0}}x_{j}^{\mu}x_{j'}^{\nu}=\sigma^{-2}(\delta_{jj'}+\sum_{m}U_{mj}U_{mj'})
\end{equation}
so we have 
\begin{equation}
\langle{\bf w}{\bf w}^{\top}\rangle_{{\bf w}}=\sigma^{2}(I+{\bf U}{\bf U}^{\top})^{-1}=\sigma^{2}{\bf I}-\sigma^{2}{\bf U}({\bf I}+{\bf U}^{\top}{\bf U})^{-1}{\bf U}^{\top}
\end{equation}
\begin{align}
\langle\sum_{jj'}U_{mj}U_{nj'}\langle{\bf w}{\bf w}^{\top}\rangle_{{\bf w}}^{jj'}\rangle_{{\bf t}} & =\langle\sigma^{2}{\bf U}^{\top}{\bf U}-\sigma^{2}{\bf U}^{\top}{\bf U}({\bf I}+{\bf U}^{\top}{\bf U})^{-1}{\bf U}^{\top}{\bf U}\rangle_{{\bf t}}
\end{align}
as we defined in SI Eq.$\text{\ref{eq:hmn}}$, $\mathcal{H}={\bf U}^{\top}{\bf U}$,
so we have 
\begin{align}
\langle\frac{1}{N}\sum_{i}a_{m,i}a_{n,i}\rangle & =\sigma^{2}{\bf I}-\langle\sum_{jj'}U_{mj}U_{nj'}\langle{\bf w}{\bf w}^{\top}\rangle_{{\bf w}}^{jj'}\rangle_{{\bf t}}=\sigma^{2}(I+\mathcal{H})^{-1}={\bf U}_{1}
\end{align}

\subsection{\label{subsec:Multiple-hidden-layers}Multiple hidden layers}

We can extend our above calculation for GGDLNs to multiple layers
by successive backward integration of weights. In this section, we
outline the derivation for a two hidden-layer network, and provide
the Hamiltonian for networks of general $L$. 

Starting from the partition function after integration of ${\bf a}$,
given by SI Eq.$\text{\ref{eq:z(W)}}$, for a network with 2 hidden
layers we have $x_{L,i}^{\mu}=x_{2,i}^{\mu}=\frac{1}{\sqrt{N_{0}M}}\sum_{mj}W_{2,ij}^{m}x_{1,j}^{\mu}g_{m}({\bf x}^{\mu}).$
We can again integrate ${\bf W}_{2}$ and obtain an equation of the
same form as SI Eq.$\text{\ref{eq:g(t)} }$, 

\begin{align}
Z & =\int\Pi_{\mu=1}^{P}dt_{\mu}\int d{\bf W}_{1}\exp[i{\bf t}^{\top}{\bf Y}+NG({\bf t})-\frac{1}{2\sigma^{2}}\mathrm{Tr}({\bf W}_{1}^{\top}{\bf W}_{1})]\\
G({\bf t}) & =\log\langle\exp-\frac{1}{2N}{\bf t}^{\top}{\bf K}_{2}^{{\bf w}}{\bf t}\rangle_{{\bf w}}\label{eq:g(t)-1}
\end{align}
now the average is with respect to a \textit{single} $N_{0}M$-dimensional
weight vector ${\bf w}_{2,i}$ with i.i.d. $\mathcal{N}(0,\sigma^{2})$
components, and $K_{2}^{{\bf w},\mu\nu}=[\frac{\sigma^{2}}{M}{\bf g}({\bf x}^{\mu})^{\top}{\bf g}({\bf x}^{\nu})]\frac{1}{N_{0}M}\sum_{jj',mm'}x_{1,j}^{\mu}g_{m}({\bf x}^{\mu})w_{2,ij}^{m}w_{2,ij'}^{m'}x_{1,j'}^{\nu}g_{m'}({\bf x}^{\nu})$.
The term in the exponent is therefore still quadratic in ${\bf w}_{2,i}$
with 
\begin{align}
\langle\exp-\frac{1}{2N}{\bf t}^{\top}{\bf K}_{2}^{{\bf w}}{\bf t}\rangle_{{\bf w}} & =\int d{\bf w}_{2,i}\exp(-\frac{1}{2}{\bf w}_{2,i}^{\top}{\bf M}{\bf w}_{2,i})\\
M_{mj,nk} & =\sigma^{-2}\delta_{jk}\delta_{mn}+\frac{1}{N}\sum_{\mu\nu}t_{\mu}t_{\nu}[\frac{\sigma^{2}}{M}{\bf g}({\bf x}^{\mu})^{\top}{\bf g}({\bf x}^{\nu})]\frac{1}{N_{0}}x_{1,j}^{\mu}g_{m}({\bf x}^{\mu})x_{1,k}^{\nu}g_{n}({\bf x}^{\nu})\label{eq:M-1}
\end{align}
performing this averaging again yields $G({\bf t})=-\frac{1}{2}\log\det(I+\mathcal{H}_{2}^{1})$,
but with
\begin{equation}
\mathcal{H}_{2}^{1,mn}=\frac{1}{N}\sum_{\mu\nu}t_{\mu}t_{\nu}\frac{\sigma^{2}}{M}g_{m}({\bf x}^{\mu})g_{n}({\bf x}^{\nu})K_{1}^{\mu\nu}\label{eq:hmn-1}
\end{equation}
where $K_{1}^{\mu\nu}$ as defined in main text is given by $K_{1}^{\mu\nu}=(\frac{\sigma^{2}}{M}{\bf g}({\bf x}^{\mu})^{\top}{\bf g}({\bf x}^{\nu}))(\frac{1}{N}{\bf x}_{1}^{\mu\top}{\bf x}_{1}^{\nu})$.
To integrate over $t$, we enforce the identity SI Eq.$\text{\ref{eq:hmn-1}}$,
by Fourier representation of the $\delta$-function, introducing the
auxiliary variable ${\bf U}_{2}^{1}$. 
\begin{multline}
Z=\int d\bm{W}_{1}\int d{\bf U}_{1}^{2}\int d\mathcal{H}_{1}^{2}\int d{\bf t}\exp[i{\bf t}^{\top}{\bf Y}-\frac{N}{2}\log\det({\bf I}+\mathcal{H}_{2}^{1})+\frac{N}{2\sigma^{2}}\mathrm{Tr}({\bf U}_{2}^{1}\mathcal{H}_{2}^{1})\\
-\frac{1}{2}{\bf t}^{\top}\tilde{{\bf K}}_{2}^{1}{\bf t}-\frac{1}{2\sigma^{2}}\mathrm{Tr}({\bf W}_{1}^{\top}{\bf W}_{1})]
\end{multline}
\begin{equation}
\tilde{K}_{2}^{1,\mu\nu}=\frac{1}{NM}\sum_{mn}\hat{H}_{2}^{1,mn}g_{m}({\bf x}^{\mu})^{\top}g_{n}({\bf x}^{\nu})K_{1}^{\mu\nu}
\end{equation}
Now we can integrate ${\bf W}_{1}$ and obtain 
\begin{equation}
Z=\int d{\bf U}_{2}^{1}\int d\mathcal{H}_{2}^{1}\int d{\bf t}\exp[i{\bf t}^{\top}{\bf Y}-\frac{N}{2}\log\det({\bf I}+\mathcal{H}_{2}^{1})+\frac{N}{2\sigma^{2}}\mathrm{Tr}({\bf U}_{2}^{1}\mathcal{H}_{2}^{1})+NG({\bf t})]
\end{equation}
\begin{equation}
G({\bf t})=\log\langle\exp-\frac{1}{2N}{\bf t}^{\top}\tilde{{\bf K}}_{2}^{1,{\bf w}}{\bf t}\rangle_{{\bf w}}
\end{equation}
where $\tilde{K}_{2}^{1,w,\mu\nu}=\frac{1}{NM}({\bf g}^{\top}({\bf x}^{\mu}){\bf U}_{2}^{1}{\bf g}({\bf x}^{\nu}))\circ(\frac{\sigma^{2}}{M}{\bf g}^{\top}({\bf x}^{\mu}){\bf g}({\bf x}^{\nu}))\circ(\frac{1}{N_{0}}{\bf x}^{\mu\top}{\bf w}_{1,i}{\bf w}_{1,i}^{\top}{\bf x}^{\nu})$,
and the average is w.r.t. a single $N_{0}$-dimensional weight vector
${\bf w}_{1,i}$ with $\mathcal{N}(0,\sigma^{2})$ components. Therefore
the term in the exponent is quadratic in ${\bf w}_{1,i}$ with 
\begin{align}
\langle\exp-\frac{1}{2N}{\bf t}^{\top}\tilde{{\bf K}}_{2}^{1,{\bf w}}{\bf t}\rangle_{{\bf w}} & =\int d{\bf w}_{1,i}\exp(-\frac{1}{2}{\bf w}_{1,i}^{\top}{\bf M}{\bf w}_{1,i})\\
M_{jk} & =\sigma^{-2}\delta_{jk}+\frac{1}{N}\sum_{\mu\nu}t_{\mu}t_{\nu}[\frac{\sigma^{2}}{M}{\bf g}({\bf x}^{\mu})^{\top}{\bf g}({\bf x}^{\nu})][\frac{1}{M}{\bf g}({\bf x}^{\mu})^{\top}{\bf U}_{2}^{1}{\bf g}({\bf x}^{\nu})]\frac{1}{N_{0}}x_{j}^{\mu}x_{k}^{\nu}\label{eq:M-1-1}
\end{align}
Performing the integral introduces another order parameter and yields
$G({\bf t})=-\frac{1}{2}\log\det({\bf I}+\mathcal{H}_{2}^{2})$, with
\begin{equation}
\mathcal{H}_{2}^{2,mm',nn'}=\frac{1}{N}\sum_{\mu\nu}t_{\mu}t_{\nu}[\frac{\sigma^{2}}{M}g_{m}({\bf x}^{\mu})g_{n}({\bf x}^{\nu})][\frac{1}{M}\tilde{g}_{m'}^{2,1}({\bf x}^{\mu})\tilde{g}_{n'}^{2,1}({\bf x}^{\mu})]K_{0}^{\mu\nu}\label{eq:hmn2}
\end{equation}
\begin{equation}
\tilde{g}_{m}^{2,1}({\bf x}^{\mu})=\sum_{m'}[{\bf U}_{2}^{1}]_{mm'}^{1/2}g_{m'}({\bf x}^{\mu})
\end{equation}
We again enforce the identity SI Eq.$\text{\ref{eq:hmn2}}$, and introduce
the auxiliary variable ${\bf U}_{2}^{2}$. 
\begin{align}
Z & =\int d{\bf U}_{2}^{1}\int d\mathcal{H}_{2}^{1}\int d{\bf U}_{2}^{2}\int d\mathcal{H}_{2}^{2}\int d{\bf t}\exp[i{\bf t}^{\top}{\bf Y}-\frac{N}{2}\log\det({\bf I}+\mathcal{H}_{1})+\frac{N}{2\sigma^{2}}\mathrm{Tr}({\bf U}_{1}\mathcal{H}_{1})\\
 & -\frac{N}{2}\log\det({\bf I}+\mathcal{H}_{2}^{2})+\frac{N}{2\sigma^{2}}\mathrm{Tr}({\bf U}_{2}^{2}\mathcal{H}_{2}^{2})-\frac{1}{2}{\bf t}^{\top}\tilde{{\bf K}}_{2}{\bf t}]
\end{align}
with $\tilde{K}_{2}^{\mu\nu}=\sum_{mn,m'n'}\frac{1}{M^{2}}g_{m}({\bf x}^{\mu})\tilde{g}_{m'}^{2,1}({\bf x}^{\mu})U_{2}^{2,mm',nn'}g_{n}({\bf x}^{\nu})\tilde{g}_{n'}^{2,1}({\bf x}^{\nu}))K_{0}^{\mu\nu}$.
Integrate over ${\bf t}$, 
\begin{multline}
Z=\int d\mathcal{H}_{2}^{2}\int d\mathcal{H}_{2}^{1}\int d{\bf U}_{2}^{1}\int d{\bf U}_{2}^{2}\exp[-\frac{N}{2}\log\det({\bf I}+\mathcal{H}_{1})+\frac{N}{2\sigma^{2}}\mathrm{Tr}({\bf U}_{1}\mathcal{H}_{1})\\
-\frac{N}{2}\log\det({\bf I}+\mathcal{H}_{2}^{2})+\frac{N}{2\sigma^{2}}\mathrm{Tr}({\bf U}_{2}^{2}\mathcal{H}_{2}^{2})-\frac{1}{2}{\bf Y}^{\top}\tilde{{\bf K}}_{2}^{-1}{\bf Y}-\frac{1}{2}\log\det\tilde{{\bf K}}_{2}]\label{eq:partition-integrated-1}
\end{multline}
The saddle-point equations can be therefore obtained by taking derivatives
w.r.t. $\mathcal{H}_{2}^{1}$ , $\mathcal{H}_{2}^{2}$, ${\bf U}_{2}^{1}$
and ${\bf U}_{2}^{2}$. One of the saddle-point equations (by taking
derivative w.r.t. $\mathcal{H}_{2}^{1}$ and $\mathcal{H}_{2}^{2}$)
yields ${\bf U}_{2}^{1}=\sigma^{2}({\bf I}+\mathcal{H}_{2}^{1})^{-1}$
and ${\bf U}_{2}^{2}=\sigma^{2}({\bf I}+\mathcal{H}_{2}^{2})^{-1}$.
Plugging into SI Eq.$\text{\ref{eq:partition-integrated-1}}$, we
have $Z=\int d{\bf U}_{2}^{1}\int d{\bf U}_{2}^{2}\exp(-H_{2})$ with
effective hamiltonian
\begin{align}
H_{2} & =\frac{1}{2}{\bf Y}^{\top}\tilde{{\bf K}}_{2}^{-1}{\bf Y}+\frac{1}{2}\log\det(\tilde{{\bf K}}_{2})-\frac{N}{2}\sum_{l=1}^{2}\log\det{\bf U}_{2}^{l}+\frac{1}{2\sigma^{2}}N\sum_{l=1}^{2}\mathrm{Tr}({\bf U}_{2}^{l})\\
\,\tilde{K}_{2}^{\mu\nu} & =\sum_{mn,m'n'}\frac{1}{M^{2}}\tilde{g}_{m,m'}^{2,2}({\bf x}^{\mu})\tilde{g}_{n,n'}^{2,2}({\bf x}^{\nu})K_{0}^{\mu\nu},\ \tilde{g}_{m,m'}^{2,2}({\bf x}^{\mu})=\sum_{nn'}[{\bf U}_{2}^{2}]_{mm',nn'}^{1/2}g_{n}({\bf x}^{\mu})\tilde{g}_{n'}^{2,1}({\bf x}^{\mu})
\end{align}
Now we can solve the saddlepoint equation by taking derivative w.r.t.
${\bf U}_{2}^{1}$ and ${\bf U}_{2}^{2}$. Note that for $L=2$, we
have 2 matrix parameters of size $M\times M$ and $M^{2}\times M^{2}$
renormalizing the kernel. 

We can iteratively perform the integration for networks of arbitrary
$L$. The partition function for networks of general $L$ is given
by $Z=\int\Pi d{\bf U}_{L}^{l}\exp(-H_{L})$ with effective Hamiltonian
\begin{equation}
H_{L}=\frac{1}{2}{\bf Y}^{\top}\tilde{{\bf K}}_{L}^{-1}{\bf Y}+\frac{1}{2}\log\det(\tilde{{\bf K}}_{L})-\frac{N}{2}\sum_{l=1}^{L}\log\det{\bf U}_{L}^{l}+\frac{1}{2\sigma^{2}}N\sum_{l=1}^{L}\mathrm{Tr}({\bf U}_{L}^{l})
\end{equation}
\begin{equation}
\tilde{K}_{L}^{\mu\nu}=\sum_{m_{1},\cdots,m_{L}}\sum_{n_{1},\cdots,n_{L}}\frac{1}{M^{L}}\tilde{g}_{m_{1},\cdots,m_{L}}^{L}({\bf x}^{\mu})\tilde{g}_{n_{1},\cdots,n_{L}}^{L}({\bf x}^{\nu})K_{0}^{\mu\nu}\label{eq:kernelmatrix}
\end{equation}
\begin{equation}
\tilde{g}_{m_{1},\cdots,m_{L}}^{L,l}({\bf x}^{\mu})=\sum_{n_{1,}\cdots,n_{l}}[{\bf U}_{L}^{l}]_{m_{1},\cdots,m_{l},n_{1},\cdots,n_{l}}^{1/2}\tilde{g}_{n_{1},\cdots,n_{l-1}}^{L,l-1}({\bf x}^{\mu})g_{n_{l}}({\bf x}^{\mu})\label{eq:iterateg}
\end{equation}

We now have $L$ matrix order parameters ${\bf U}_{L}^{l}\in\mathbb{R}^{M^{l}\times M^{l}}$($l=1,\cdots,L$).
Note that the size of the order parameter matrix grows exponentially
with $L$, limiting the application of our theory in very deep networks
in practice.

\subsection{Generalization}
\label{subsec:Generalization}
The mean-squared generalization error depends only on the mean and
variance of the predictor, which can be computed using the generating
function
\begin{multline}
Z(t_{P+1})=\int d\Theta\exp[-\frac{1}{2T}\sum_{\mu=1}^{P}(\frac{1}{\sqrt{NM}}\sum_{i=1}^{N}\sum_{m=1}^{M}a_{m,i}x_{L,i}^{\mu}g_{m}({\bf x}^{\mu})-Y^{\mu})^{2}\\
+it_{P+1}\frac{1}{\sqrt{N}}\sum_{i=1}^{N}\sum_{m=1}^{M}a_{m,i}x_{L,i}g_{m}({\bf x})-\frac{1}{2\sigma^{2}}\Theta^{\top}\Theta]
\end{multline}
where ${\bf x}$ is an arbitrary new point. The statistics of the
predictor are given by 
\begin{equation}
\langle f({\bf x})\rangle=\partial_{it_{P+1}}\log Z\lvert_{t_{P+1}=0}
\end{equation}
\begin{equation}
\langle\delta f({\bf x})^{2}\rangle=\partial_{it_{P+1}}^{2}\log Z\lvert_{t_{P+1}=0}
\end{equation}
The integral can be performed similarly as in Section $\text{\ref{subsec:Kernel-shape-renormalization}}$
and $\text{\ref{subsec:Multiple-hidden-layers}}$, after integrating
all weights, we obtain 
\begin{multline}
Z(t_{P+1})=\int\Pi d{\bf U}_{L}^{l}\exp[-\frac{N}{2}\sum_{l=1}^{L}\log\det{\bf U}_{L}^{l}+\frac{1}{2\sigma^{2}}N\sum_{l=1}^{L}\mathrm{Tr}({\bf U}_{L}^{l})\\
+\frac{1}{2}(i{\bf Y}+t_{P+1}^{\top}\tilde{{\bf k}}_{L}({\bf x}))^{\top}\tilde{{\bf K}}_{L}^{-1}(i{\bf Y}+t_{P+1}^{\top}\tilde{{\bf k}}_{L}({\bf x}))-\frac{1}{2}\log\det\tilde{{\bf K}}_{L}-\frac{1}{2}t_{P+1}^{\top}\tilde{K}_{L}({\bf x},{\bf x})t_{P+1}]
\end{multline}
Here 
\begin{equation}
\begin{split}
\tilde{K}_{L}({\bf x},{\bf y})&=\sum_{m_{1},\cdots,m_{L}}\sum_{n_{1},\cdots,n_{L}}\frac{1}{M^{L}}\tilde{g}_{m_{1},\cdots,m_{L}}^{L}({\bf x})\tilde{g}_{n_{1},\cdots,n_{L}}^{L}({\bf y})K_{0}({\bf x},{\bf y})\\
\tilde{g}_{m_{1},\cdots,m_{L}}^{L,l}({\bf x})&=\sum_{n_{1,}\cdots,n_{l}}[{\bf U}_{L}^{l}]_{m_{1},\cdots,m_{l},n_{1},\cdots,n_{l}}^{1/2}\tilde{g}_{n_{1},\cdots,n_{l-1}}^{L,l-1}({\bf x})g_{n_{l}}({\bf x})
\end{split}
\end{equation}
$\tilde{{\bf K}}_{L}$ denotes the $P\times P$ kernel matrix evaluated
on the training data, as given by Eqs.$\text{\ref{eq:kernelmatrix},\ref{eq:iterateg}}$,
and $\tilde{{\bf k}}_{L}({\bf x})$ is a $P$-dimensional vector with
$\tilde{k}_{L}^{\mu}({\bf x})=\tilde{K}_{L}({\bf x},{\bf x}^{\mu})$.
Differentiating $\log Z$, we obtain
\begin{equation}
\langle f({\bf x})\rangle=\partial_{it_{P+1}}\log Z\lvert_{t_{P+1}=0}=\tilde{{\bf k}}_{L}({\bf x})^{\top}\tilde{{\bf K}}_{L}^{-1}{\bf Y}
\end{equation}
\begin{equation}
\langle\delta f({\bf x})^{2}\rangle=\partial_{it_{P+1}}^{2}\log Z\lvert_{t_{P+1}=0}=\tilde{{\bf K}}_{L}({\bf x},{\bf x})-\tilde{{\bf k}}_{L}({\bf x})^{\top}\tilde{{\bf K}}_{L}^{-1}\tilde{{\bf k}}_{L}({\bf x})
\end{equation}

For some simulation in the main text, Figs.$\text{\ref{fig:sigdependence},\ref{fig:pretrainedgatings},\ref{fig:permutedmnist}}$ and SI Figs.$\text{\ref{fig:Simulation-with-gd1},\ref{fig:Simulation-with-gd2}}$,
we considered the classification error, it is obtained by approximating
the predictor on each data point $x$ to be Gaussian with mean $\langle f({\bf x})\rangle$
and variance $\langle\delta f({\bf x})^{2}\rangle$, so that the error
rate is given by 
\begin{equation}
error\ rate=(y({\bf x})+1)/2-y({\bf x})\frac{1}{2}\mathrm{erfc}(\frac{-\langle f({\bf x})\rangle}{\sqrt{2\langle\delta f({\bf x})^{2}\rangle}})\label{eq:errorrate}
\end{equation}

Here $y({\bf x})\in\{\pm1\}$. 

\section{\label{sec:Additional-numerical-results}Additional numerical results}

\subsection{\label{subsec:Double-descent-with}Double descent with $M$ for finite
width GGDLNs}

\begin{SCfigure}[][h]

\includegraphics[width=0.6\textwidth]{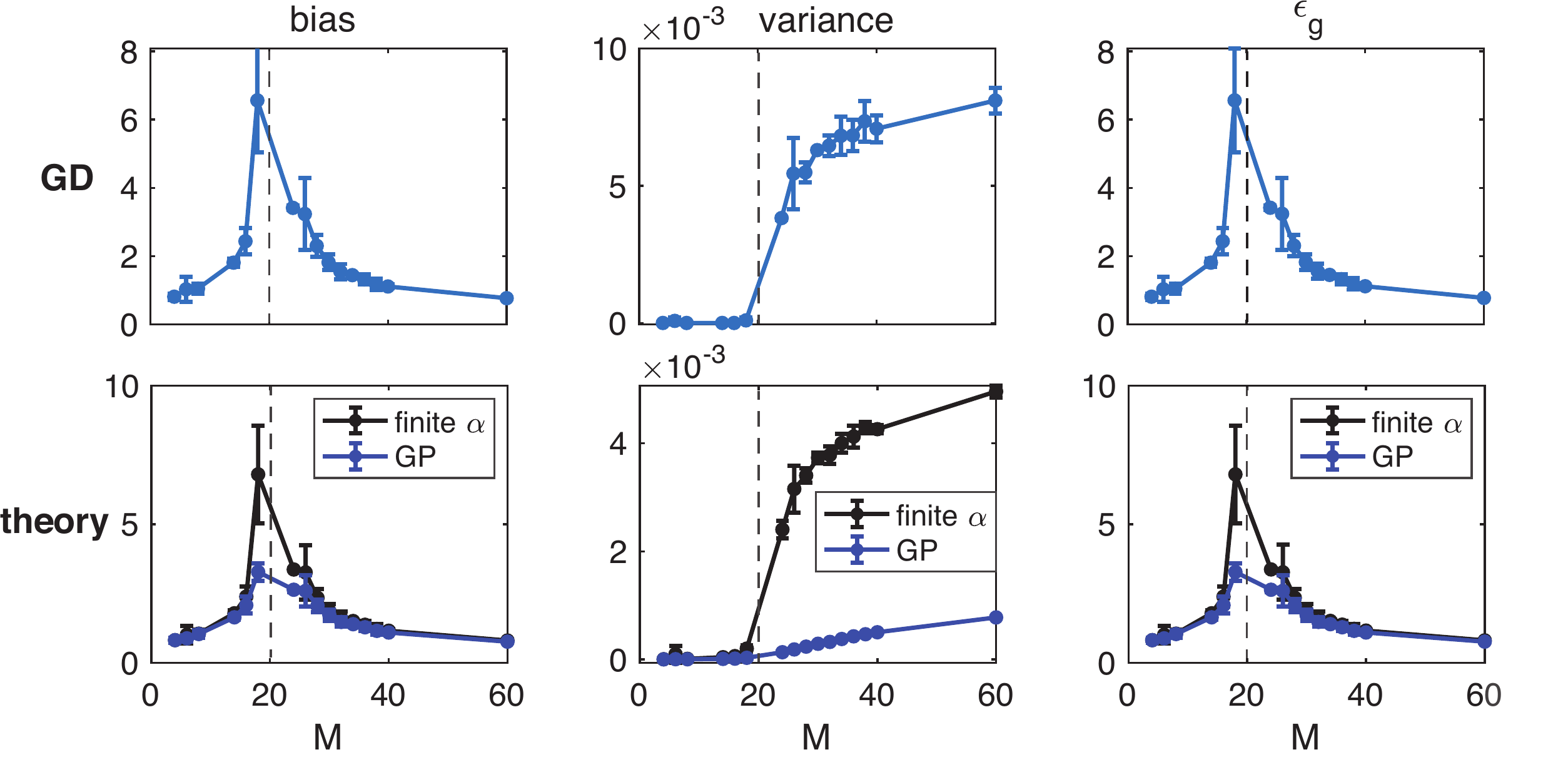}\caption{Double descent with $M$ for finite width GGDLNs. Bias, variance and generalization error of networks with a single hidden layer at finite width evaluated on MNIST. Bias and generalization error diverges, and variance becomes nonzero at the network capacity (black dashed line). Finite width theory (bottom black line) agrees well with GD dynamics (top) qualitatively, and is more accurate compared to GP (bottom blue line).}
\label{fig:finitewidthcapacity}
\end{SCfigure}

In Fig.$\text{\ref{fig:Globally-gated-deep} }$ we showed the double
descent phenomenon for GGDLNs in the GP limit. Here we show that the
singularity of the kernel at the interpolation threshold holds even
for finite width networks, and similar diverging bias and vanishing
variance are seen in the finite width theory (SI Fig.$\text{\ref{fig:finitewidthcapacity}}$
bottom) with kernel shape renormalization, and are confirmed by simulation
of networks trained with GD (SI Fig.$\text{\ref{fig:finitewidthcapacity}}$
top). Our theory with renormalized kernel agrees better with the simulation
with GD dynamics compared to the theory in the GP limit. 

\subsection{\label{subsec:Langevin-dynamics}Langevin dynamics}

Throughout the main text, we compare our theoretical prediction with
simulation using GD dynamics, which is more commonly used in practice.
Since our theory does not directly describes GD dynamics but the properties
of the posterior distribution of the network weights induced at equilibrium
by Langevin dynamics with the MSE cost function and the Gaussian prior,
qualitative agreement between the theory and the GD simulations is
already a remarkable result. Here we compare our theory with simulation
with the corresponding Langevin dynamics, given by 
\begin{equation}
\text{\ensuremath{\Delta\Theta=-\epsilon\partial_{\Theta}E}}+\sqrt{2\epsilon T}\eta
\end{equation}
with 
\begin{equation}
E=\frac{1}{2}\sum_{\mu=1}^{P}(f({\bf x}^{\mu},\Theta)-Y^{\mu})^{2}+\frac{T}{2\sigma^{2}}\Theta^{\top}\Theta
\end{equation}
as the loss function, which equates the exponent in the posterior
distribution Eq.$\text{\ref{eq:posterior}}$. Here $\epsilon$ denotes
the step size, and $\eta$ denotes standard Gaussian white noise.
We see that the theory and Langevin dynamics simulation agrees \textit{quantitatively}
accurate, and the simulation dots lie right on top of the line of
our theoretical prediction. The example is the same task and same
parameter as in SI Fig.$\text{\ref{fig:Dependence-of-generalization}}$. 

\begin{figure}[h]
\centering

\includegraphics[width=0.8\textwidth]{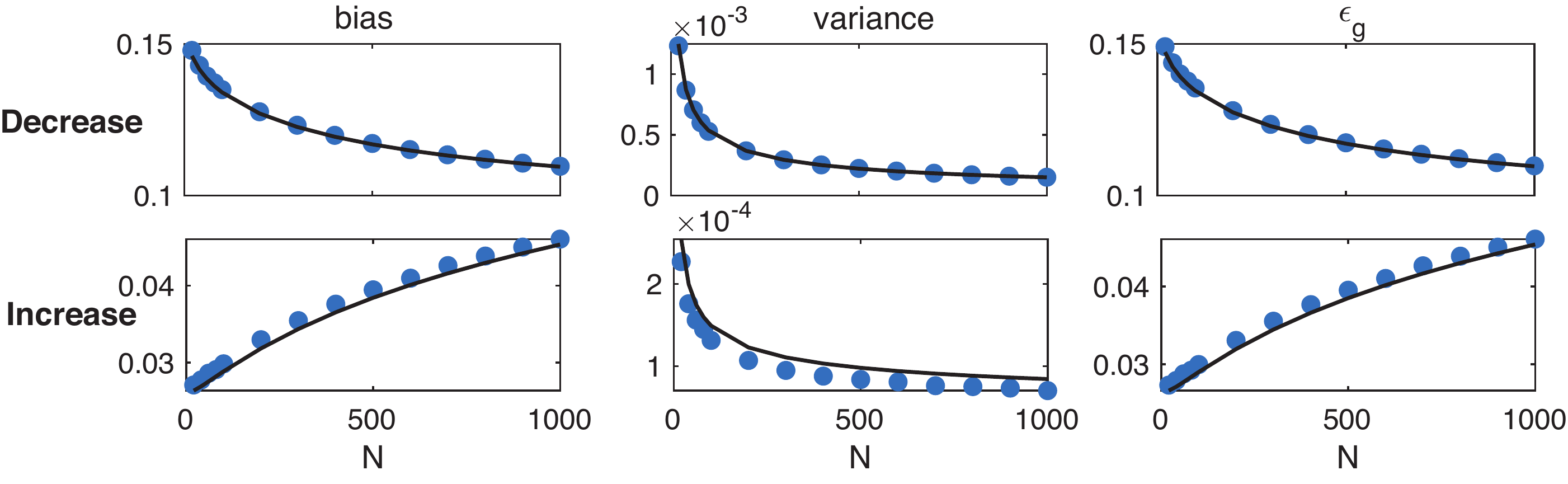}

\caption{\label{fig:Simulation-with-Langevin}Simulation with Langevin dynamics.
Dependence of generalization on network width, with the same task
and parameters as in Fig.$\text{\ref{fig:Globally-gated-deep}}$.
Black lines: Theory. Blue dots: Simulation with gradient-based Langevin
dynamics. Our theoretical prediction of generalization perperties
obtained from running Langevin dynamics is quantitatively accurate.}
\end{figure}

\subsection{\label{subsec:Gradient-descent-dynamics}Gradient-descent dynamics }

\begin{figure}[h]
\centering

\includegraphics[width=0.8\textwidth]{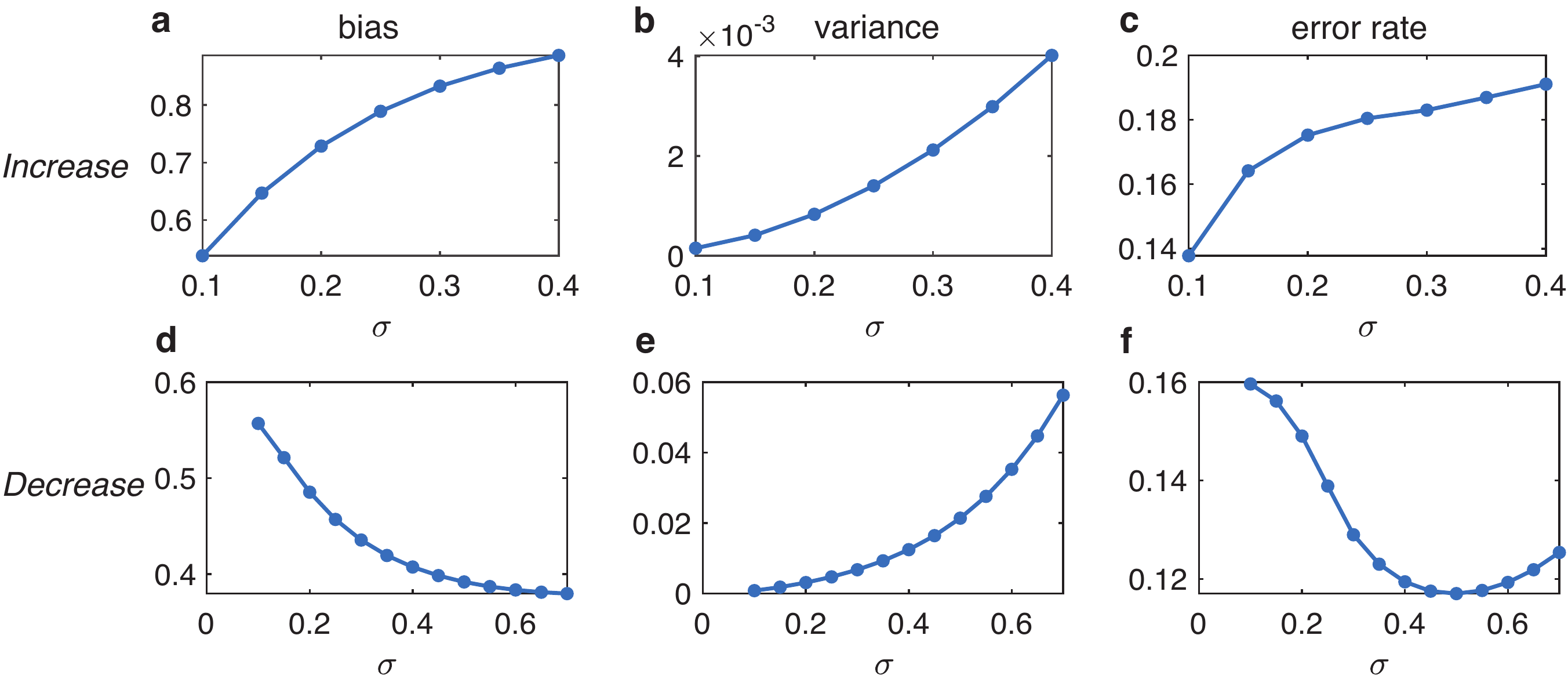}

\caption{\label{fig:Simulation-with-gd1}Simulation with GD dynamics for dependence
of generalization on weight regularization strength $\sigma$, with
the same task and parameters as in main text Fig.$\text{\ref{fig:sigdependence}}$.
The simulation exhibits similar generalization behavior as our theory
predicts in the two parameter regimes. (\textbf{a}-\textbf{c}) Bias
(a), variance (b), and error rate (c) increases as a function of $\sigma$
as predicted by theory. (\textbf{d}-\textbf{f}) Bias (d) decreases
as a function of $\sigma$ while variance (e) increases, resulting
in an optimal $\sigma$ where the error rate is at its minimum (f).}
\end{figure}

In Fig\@.$\text{\ref{fig:sigdependence}}$ and Fig.$\text{\ref{fig:pretrainedgatings}}$
in the main text, we present only the prediction of our theory of
how generalization performance depends on the regularization strength
$\sigma$ and the number of gatings $M$. Here in SI Figs.\text{\ref{fig:Simulation-with-gd1},\ref{fig:Simulation-with-gd2}} we show our results
obtained from GD dynamics (detailed in Appendix $\text{\ref{subsec:Gradient-descent-numerics}}$)
and show that they agree well with the theoretical predictions, exhibiting
qualitatively similar behavior with the theoretical results in the
corresponding parameter regimes.

\begin{figure}[h]
\centering

\includegraphics[width=0.8\textwidth]{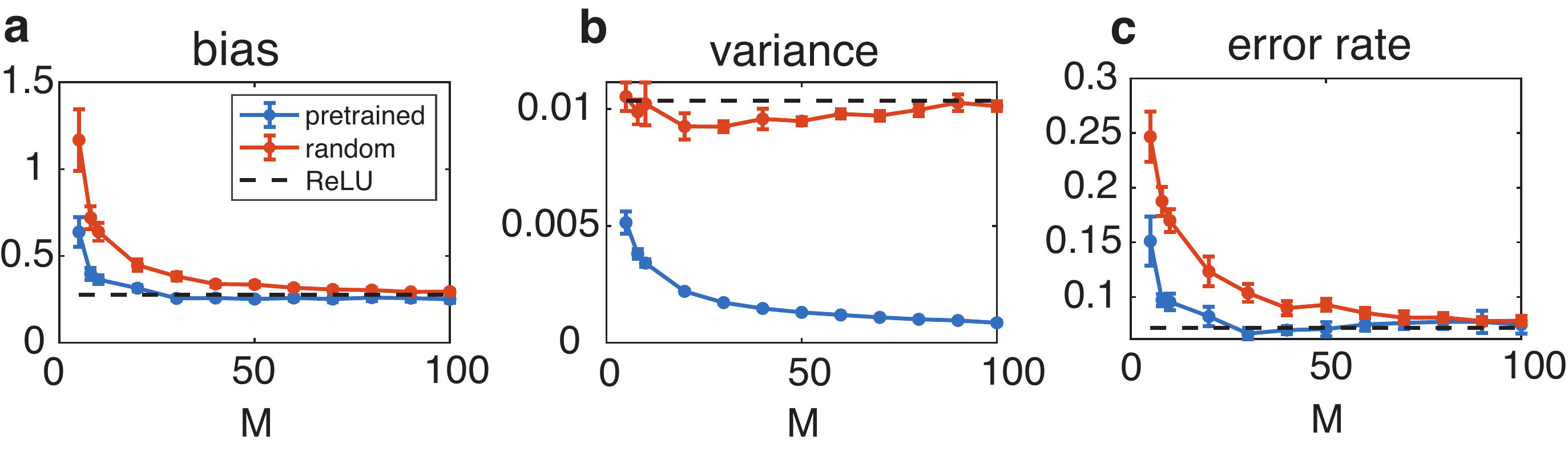}

\caption{\label{fig:Simulation-with-gd2}Simulation with GD dynamics for dependence
of generalization on M for GGDLNs trained on MNIST dataset, with the
same task and parameters as in main text Fig. $\text{\ref{fig:pretrainedgatings}}$.
Bias (a), variance (b) and error rate (c) as a function of M for random
(red lines) and pretrained gatings (blue lines). Performance of GGDLNs
improves and approaches ReLU network (black dashed lines) for sufficiently
large M, and improves faster for pretrained gatings compared to random
gatings. All these qualitative properties agree with theoretical predictions.}
\end{figure}

\subsection{\label{subsec:Deep-GP-kernel}Deep GP kernel of GGDLNs}

The normalized GP kernel of GGDLNs is given by 
\begin{equation}
\mathcal{K}_L({\bf x},{\bf y})=\cos({\bf g}({\bf x}),{\bf g}({\bf y}))^{L}\cos({\bf x},{\bf y})
\end{equation}
the exact shape of which depends on the specific choice of gatings.
However, as discussed in the main text, in the limit of \textit{random
gatings }where with zero threshold $g_{m}({\bf x})=\Theta(\frac{1}{\sqrt{N_{0}}}{\bf V}^{\top}{\bf x})$,
${\bf V}\sim\mathcal{N}(0,1)$, and the number of gatings $M\rightarrow\infty$.
We have $\cos({\bf g}({\bf x}),{\bf g}({\bf y}))\rightarrow\frac{\pi-\theta}{\pi}$,
as derived in \cite{cho2009kernel}, and 
\begin{equation}
\mathcal{K}_{L}({\bf x},{\bf y})=(\frac{\pi-\theta}{\pi})^{L}\cos\theta\label{eq:kgp}
\end{equation}

\begin{figure}[h]
\centering

\includegraphics[width=1\textwidth]{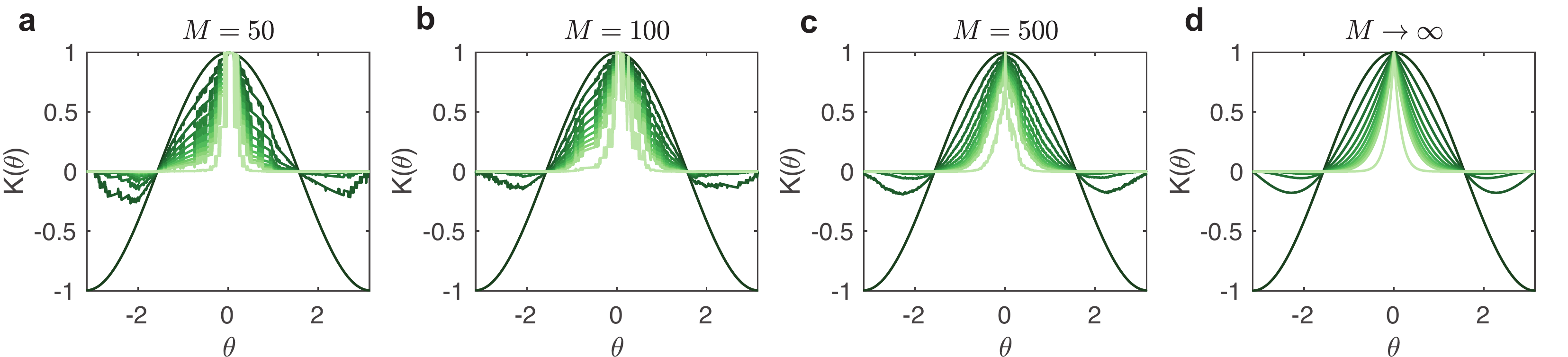}

\caption{\label{fig:GP-kernel-of}GP kernel of GGDLNs for \textit{random gatings}
with zero threshold as a function of input angles for $M=50$ (a),
$M=100$ (b), $M=500$ (c) calculated numerically, and for $M\rightarrow\infty$
(d) calculated analytically using SI Eq.$\text{\ref{eq:kgp}}$ for
$L=0-10$, lighter colors are for larger $L$. As $L$ increases the
kernel gradually shrinks to
0 for any $\theta\protect\neq0$, exhibiting a 'flattening' effect where input information is gradually lost.}
\end{figure}

For finite $M$, we numerically calculate the normalized GP kernel
on inputs with different angles $\theta$ between them. We generate
inputs with different angles between them by constraining them roughly
in a 2-D subspace ${\bf x}(\theta)=[\cos\theta,\sin\theta,\eta_{1},\cdots,\eta_{N_{0}-2}]$,
where $\eta_{1},\cdots,\eta_{N_{0}-2}\sim\mathcal{N}(0,\sigma_{0}^{2})$
with $\sigma_{0}=0.005$. Then we numerically compute $\mathcal{K}_{L}(\theta)=\cos(g({\bf x}(0)),g({\bf x}(\theta)))^{L}\cos({\bf x}(0),{\bf x}(\theta))$.
For $M\rightarrow\infty$, we plot the analytical expression SI Eq.$\text{\ref{eq:kgp}}$
in SI Fig.$\text{\ref{fig:GP-kernel-of}}$(d). 

\subsection{Renormalized kernel and performance of GGDLNs with multiple hidden
layers}

\begin{figure}[h]
\centering

\includegraphics[width=1\textwidth]{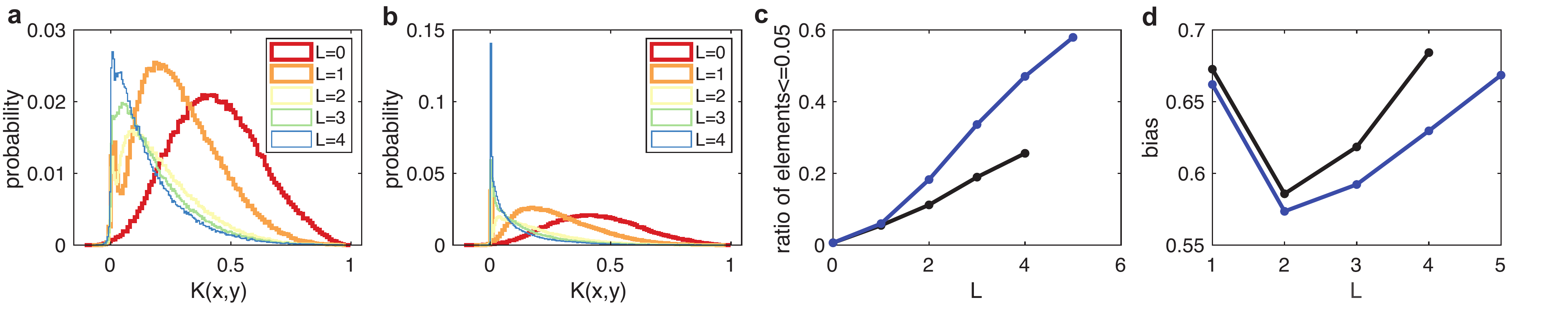}

\caption{\label{fig:Shape-renormalization-slows-1}Shape renormalization slows
down flattening of kernels in deep networks trained on MNIST. (\textbf{a}-\textbf{c})
Similar to Fig\@.$\text{\ref{fig:Shape-renormalization-slows}}$,
shape renormalization slows down the flattening of the kernel and
prevents the elements from quickly shrinking to 0. (\textbf{d}) The
bias contribution to the generalization first decreases then increases
as a function of $L$. Finite width network with renormalized kernel
(black line) performs worse than the GP (blue line) for $L>1$ in
this parameter regime. See Appendix $\text{\ref{sec:Detailed-parameters}}$
for detailed parameters. }
\end{figure}

In the main text, we showed that the effect of kernel renormalization
slows down the 'flattening' of the GP kernel. However, it is not clear
what effect it has on the generalization error (especially bias),
and whether generalization improves or degrades depends on the specific
parameters. In the main text, we showed an example where kernel renormalization
is beneficial for generalization, here we show another example where
kernel renormalization still mitigates the flattening of the GP kernel
(SI Fig.$\text{\ref{fig:Shape-renormalization-slows-1}}$(a-c)), but
results in a worse generalization (bias) compared to networks in the
GP limit (SI Fig.$\text{\ref{fig:Shape-renormalization-slows-1}}$(d)). 

\section{\label{sec:Detailed-parameters}Detailed parameters and setup of
simulations}

\subsection{Noisy ReLU teacher}\label{SI:noisyrelu}

The input data ${\bf x}\in\mathbb{R}^{N_0}$ is drawn from i.i.d. Gaussian distribution ${\bf x}\sim\mathcal{N}(0,{\bf I}_{N_0})$, and the test data is corrupted copies of the input data given by ${\bf x}_t = \sqrt{1-\gamma}{\bf x} + \sqrt{\gamma}{\bf \eta}, \eta\sim\mathcal{N}(0,{\bf I}_{N_0})$. The noisy ReLU teacher labels are given by $y({\bf x}) = \frac{1}{\sqrt{N_T}}{\bf a}_T \mathrm{ReLU}(\frac{1}{\sqrt{N_0}}{\bf W}_T{\bf x}) + \varepsilon \eta_T$, ${\bf W}_{T} \in\mathbb{R}^{N_T\times N_0}$, ${\bf a}_T\in\mathbb{R}^{N_T}$, $\eta_T\sim\mathcal{N}(0,1)$, both ${\bf a}_T$ and ${\bf W}_T$ are drawn from i.i.d. Gaussian. The parameters in Fig.\ref{fig:Globally-gated-deep} (b-e) are $N_0=30, P=2100, N_T=3000,\gamma=0.01,\varepsilon=0.1$. In (c-e) the number of test data points is $P_t=1000$. Results are calculated using Eq.\ref{eq:gp} in the main text.

\subsection{ReLU teacher with preferred inputs}

The input data ${\bf x}\in\mathbb{R}^{N_{0}}$ is divided into $m$
different subsets of input dimensions ${\bf x}=[{\bf x}_{1},\cdots,{\bf x}_{m}]$.
Within each subset of input dimensions, the data is arranged into
the same $n$ clusters in an $N_{0}/m$ dimensional space, ${\bf x}_{m'}=\sqrt{1-\gamma}{\bf x}_{c}^{n'}+\sqrt{\gamma}{\bf \eta}$,
$m'=1,\cdots,m$, $n'=1,\cdots,n$, $\eta\sim\mathcal{N}(0,{\bf I}_{N_0})$. Here
${\bf x}_{c}^{n'}$'s are the cluster centers. The ReLU teacher is
given by $y({\bf x})=\frac{1}{\sqrt{N_{T}}}{\bf a}_T\mathrm{ReLU}(\frac{1}{\sqrt{N_{0}}}{\bf W}_T{\bf x})$.
${\bf W}_T\sim\mathcal{N}(0,[\underbrace{\rho,\cdots,\rho}_{N_{0}/m},\underbrace{1,\cdots,1}_{N_{0}-N_{0}/m}])$. 

The student network is a GGDLN with a single hidden layer. The $M$
gatings are divided into $m$ subsets, each connecting to only one
subset of the input dimensions with standard Gaussian i.i.d. random
weights ${\bf V}\in\mathbb{R}^{N_{0}/m}$ and zero threshold. For
example, the activation of a gating unit connecting to the $m'$-th
subset is given by $g({\bf x})=\Theta(\frac{1}{\sqrt{N_{0}}}{\bf V}^{\top}{\bf x}_{m'})$. 

The parameters in Fig.$\ref{fig:Dependence-of-generalization}$(b)
and SI Fig.$\text{\ref{fig:Simulation-with-Langevin}}$ top are $N_{0}=200$,
$M=20$, $N_{t}=1000$, $P=1000$, $\gamma=0.01$, $m=10$, $n=20$,
$\rho=1$. The parameters in Fig.$\text{\ref{fig:Dependence-of-generalization}}$(c)
and SI Fig.$\text{\ref{fig:Simulation-with-Langevin}}$ bottom are
$N_{0}=100$, $M=50$, $N_{t}=1000$, $P=200$, $\gamma=0.01$, $m=5$,
$n=20$, $\rho=0.01$. 

\subsection{\label{subsec:MNIST-classification}MNIST classification}

Here we list the detailed parameters for simulation on MNIST binary
classification. Here for the random gatings the threshold $b$ is
0 for all simulations. 
\begin{enumerate}
\item In SI Fig.$\text{\ref{fig:finitewidthcapacity}}$,
the generalization behaviors are calculated on classifying even and
odd MNIST digits. We first properly normalize and center the data.
To change $N_{0}$, we project the 784 dimensional input in the MNIST
dataset onto an $N_{0}$ dimensional subspace with random weights,
and add a ReLU nonlinearity to the projected data, ${\bf x}={\rm ReLU}(\frac{1}{\sqrt{784}}{\bf W}_{0}{\bf x}_{{\rm MNIST}})$,
where ${\bf W}_{0}\in\mathbb{R}^{N_{0}\times784}$, $W_{0}\sim\mathcal{N}(0,1)$.
We again normalize and center ${\bf x}$ to have zero mean and standard
deviation 1. In SI Fig.$\text{\ref{fig:finitewidthcapacity}}$,
the parameters are $N_{0}=50$, $P=1000$, $\sigma=0.5$, $P_{t}=1000$.
The training and testing samples consist of equal amount of even and
odd digits. 
\item In Fig.$\text{\ref{fig:sigdependence}}$, the task is classifying
even and odd MNIST digits, here we directly take the normalized (standard
deviation 1) and centered (zero mean) $784$-dimensional MNIST data
as inputs. For the top panel, the parameters are $M=5$, $N=3000$,
$P=800$, $P_{t}=1000$. For the bottom panel, the parameters are
$M=100$, $N=200$, $P=300$ and $P_{t}=1000$. The training and testing
samples consist of equal amount of even and odd digits. 
\item In Fig.$\text{\ref{fig:pretrainedgatings}},$the task is classifying
even and odd MNIST digits, with the normalized and centered $784$-dimensional
MNIST data as inputs. The parameters are $N=1000$, $P=1000$, $P_{t}=1000$,
$\sigma=0.5$. The training and testing samples consist of equal amount
of even and odd digits. 
\item In Fig\@.$\text{\ref{fig:Shape-renormalization-slows}}$ and SI Fig.$\text{\ref{fig:Shape-renormalization-slows-1}}$,
the task is classifying even and odd MNIST digits, with the normalized
and centered $784$-dimensional MNIST data as inputs. In Fig.$\text{\ref{fig:Shape-renormalization-slows}}$,
the parameters are $M=6$, $N=500$, $P=600$, $P_{t}=2000$, $\sigma=1$.
In SI Fig.$\text{\ref{fig:Shape-renormalization-slows-1}}$, the parameters
are $M=8$, $N=500$, $P=600$, $P_{t}=2000$, $\sigma=1$. In both
cases the kernel elements corresponding to different digits dominate
and are close to 0 for both GP and finite width networks with different
depth, for better visualization, we show only elements corresponding
to the same digits. The training and testing samples consist of equal
amount of even and odd digits. 
\end{enumerate}

\subsection{\label{subsec:Permuted-MNIST}Permuted MNIST}

Here we list the parameters for simulation performed on permuted MNIST
with binary classification
\begin{enumerate}
\item In Fig.$\text{\ref{fig:permutedmnist}}$, the task is classification
of even and odd digits of permuted MNIST with 10 random permutations
of all 784 pixels. The training and testing samples consist of equal
amount of even and odd digits. The parameters for Fig.$\text{\ref{fig:permutedmnist}}$(a-d)
are $M=50$, $P=300$, $P_{t}=500$, $b=-2$, $\sigma=0.2$. The parameters
for Fig.$\text{\ref{fig:permutedmnist}}$ (e-f) are $M=50$, $P=300$,
$P_{t}=500$, $N=1000$, $\sigma=0.2$. 
\item In Fig.$\text{\ref{fig:Kernel-renormalization-decorrela}}$, the data
is permuted and unpermuted MNIST digits of 0's and 1's projected onto
$N_{0}$-dimensional subspace similarly as introduced in Section $\text{\ref{subsec:MNIST-classification}}$,
the data contains equal amount of unpermuted digit 0, unpermuted digit
1, permuted digit 0 and permuted digit 1. The gatings combine top-down
and bottom-up signals. For each different task, we select a random
subset of the gatings to be permitted with probability $p(permitted)=0.75$.
Among the permitted gatings, they depend on the input data through
$g_m({\bf x})=\Theta(\frac{1}{\sqrt{N_{0}}}{\bf V}_m^{\top}{\bf x})$, where the entries of ${\bf V}_m$ are i.i.d. Gaussian. The parameters are
$N_{0}=400$, $P=600$ ($300$ permuted and $300$ unpermuted), $P_{t}=500$,
$M=20$, $\sigma=1$. 
\end{enumerate}

\subsection{\label{subsec:Gradient-descent-numerics}Gradient descent numerics}

Throughout the main text we compare our theory with simulations with
GD dynamics. In the simulations, we initialize the weights from Gaussian
i.i.d. distribution with standard deviation $\sigma$ as in the Gaussian
prior in Eq.$\text{\ref{eq:posterior}}$. We then train the network
with GD dynamics with the mean squared error loss function without
the $L_{2}$ regularization term, and stop the training dynamics when
the training error is sufficiently small ($\frac{1}{P}\sum_{\mu}(f(x^{\mu},\Theta)-y^{\mu})<1e-3$).
The statistics including the mean and variance are obtained by simulating
multiple trajectories with different realizations of the initialization
weights \cite{yao2007early,advani2020high}. 

\end{document}